\documentclass[12pt,beng]{uom_eee_dissertation_casson} 
\usepackage{amsmath, amssymb, amsfonts}

\usepackage{graphicx}
\usepackage{times}

\usepackage{booktabs}
\usepackage{multirow}
\usepackage{makecell}

\usepackage{microtype}
\usepackage{setspace}

\usepackage{xcolor}

\usepackage{subcaption}

\usepackage{algorithm}
\usepackage{algorithmic}

\usepackage{amsmath}
\usepackage{float}

\begin{document}
\makeatletter
\title{VesperMUSE: Multi-Modal Fusion Vision Transformer for Image Semantic Segmentation Under Diverse Light Conditions}
\studentid{11106299}
\studentname{Raul Balmez}
\supervisor{Tingting Mu}
\makeatother

\course{Computer Science} 
\faculty{Science and Engineering}
\school{Department of Computer Science}
\submitdate{2026}
\wordcount{13696} 
\maketitle

\uomtoc
\uomlof
\uomlot

\begin{uomterms}
\begin{description}[leftmargin=!]
  \item[BN] Batch Normalization
  \item[CDF] Cumulative Distribution Function
  \item[CNN] Convolutional Neural Network
  \item[CV] Computer Vision 
  \item[DL] Deep Learning
  \item[FFN] Feed-Forward Network
  \item[FLOPs] Floating-Point Operations
  \item[GELU] Gaussian Error Linear Unit
  \item[IoU] Intersection over Union
  \item[LLIE] Low-Light Image Enhancement
  \item[LN] Layer Normalization
  \item[MAE] Mean Absolute Error
  \item[mIoU] Mean Intersection over Union
  \item[MS-SSIM] Multi-Scale Structure Similarity Index Measure
  \item[NMI] Normalized Mutual Information
  \item[ReLU] Rectified Linear Unit
  \item[SNR] Signal-to-Noise Ratio
  \item[PSNR] Peak-Signal-to-Noise Ratio
  \item[SSIM] Structural Similarity Index Measure
  \item[SOTA] State of the Art
  \item[SSR] Single-Scale Retinex
  \item[UAV] Unmanned Aerial Vehicle
  \item[ViT] Vision Transformer
  
\end{description}
\end{uomterms}

\begin{uomoriginality}
  \uomoriginalitydeclaration 
\end{uomoriginality}
\uomcopyrightstatement

\begin{uomacknowledgements}
I would like to thank my supervisor, Dr. Tingting Mu, for all the continued support.
\end{uomacknowledgements}

\begin{abstract}
    Semantic segmentation is critical for the reliable navigation and scene understanding of autonomous systems, yet its performance degrades severely in low-light environments due to noise, artifacts, colour distortion, and the loss of structural details. To address this challenge, this work introduces VesperMUSE, a unified multi-modal framework designed to achieve robust image semantic segmentation across diverse illumination conditions. The architecture addresses the light deficiency problem on multiple levels: proposing a novel physics-informed synthetic low-light adaptation algorithm, approaching Low-Light Image Enhancement (LLIE) through a thermal-guided paradigm, and finally aiming to recover the intrinsic scene details for semantic segmentation through a multi-modal fusion approach. Following enhancement, the framework dynamically integrates the primary RGB features with auxiliary thermal representations and surface normal maps highlighting geometric discontinuities and material boundaries. At the core of this integration is a highly scalable, modality-agnostic self-attention mechanism that extracts both global context and fine local features from an arbitrary number of auxiliary inputs. By combining and projecting these hierarchical multi-modal representations into a unified  component, the attention module allows the network to selectively attend to relevant cross-modal cues without requiring any architectural modifications when different sensor configurations are used. Extensive evaluations demonstrate that VesperMUSE successfully recovers missing visual cues, providing a flexible, state-of-the-art (SOTA) solution for machine perception in extreme and unstructured real-world environments.

\end{abstract}
\clearpage


\chapter{Introduction}
\newpage
\textbf{Semantic Segmentation. } The performance and goal-achievement capability of autonomous or Unmanned Aerial Vehicle (UAV) systems, which have become a preferred choice for navigating dangerous environments in recent years, rely strongly on scene understanding. High-level Computer Vision (CV) tasks such as pixel-level semantic segmentation \cite{long2015fully, badrinarayanan2017segnet} or instance-level object detection \cite{zhao2019object} have emerged as powerful tools for mapping sensor observations into information and cues that guide decision-making. The main challenge remains robust perception under varying conditions \cite{ha2017mfnet, zhou2023dbcnet}. Semantic segmentation achieves scene understanding by assigning each pixel in an image to a specific class, with classes defined during training. In this way, the entire scene is parsed, and objects can be more easily identified by their shape or distinct attributes. Due to its highly reliable approach, the applications of semantic segmentation extend beyond outdoor scene parsing to domains such as medicine \cite{ronneberger2015u}, indoor environments \cite{silberman2012indoor}, and even more challenging missions such as survivor localisation \cite{shivakumar2020pst900} in hazy or poorly lit conditions.

Usually, high-level CV tasks perform best under near-ideal conditions: consistent illumination that ensures accurate colour balance, without haze, smoke, noise, or other degradations. However, real-world environments rarely provide such conditions, therefore, robust systems must be capable of handling non-ideal sensor inputs \cite{zhang2023cmx, sun2019rtfnet}. In this study, the focus is on improving low-light semantic segmentation and understanding the steps required to recover lost detail, as well as how to compensate for features that cannot be restored \cite{liang2023explicit}.

\textbf{Low-Light Conditions. } Light-deficient environments occur as frequently as daylight scenes, making them not merely an edge case of CV systems but a domain that requires rigorous treatment \cite{liu2021benchmarking}. What makes low-light images particularly challenging is that limited illumination introduces additional issues beyond the obvious reduced exposure, such as global and local structure loss, noise, artifacts, and prominent colour distortions \cite{guo2020zero, Zhang_2019}. There are multiple approaches to improving semantic segmentation performance under such conditions, with the main categories being: enhancement-based methods \cite{EnGAN} (where a network is applied prior to segmentation to enhance the images), knowledge distillation \cite{Wu_2021} (where a student network trained on low-light images is constrained to match the internal representations of a teacher network trained on well-lit data), single-network approaches (where the feature space is analyzed using frequency-domain techniques \cite{Wang_2025} with a greater emphasis being placed on edge information), and multi-modal fusion approaches \cite{li2022rgbsegmentation, zhang2025spectralsegmentation}, in which the limited information in low-light images is compensated by incorporating additional representations.

\textbf{Thermal Imagery.} An expected observation is the reliability of thermal (infrared) sensing in low-light environments, as such representations do not depend on illumination but rather on heat irradiation, making them largely light-agnostic \cite{Kristo_2020} and capable of preserving details often lost in low-light imagery, such as clear structural boundaries and reduced artifacts \cite{Hwang_2015}. However, this modality presents several limitations, including poor separation of objects or surfaces with similar temperatures \cite{Vollmer_2017}, background interference when ambient temperatures approximate those of objects \cite{Gonz_lez_2016}, and the absence of colour information, which remains an essential cue for accurate scene parsing and object identification \cite{Sun_2021}. Despite these constraints, the practical acquisition of thermal data has become increasingly accessible. The commercial miniaturization of Long-Wave Infrared micro-bolometers has rendered these sensors both cost-effective and easy to integrate into real-world autonomous systems. In the scope of this study, the thermal modality is technically leveraged to enhance low-light RGB data by providing a high Signal-to-Noise Ratio (SNR) in environments where visible-spectrum sensors succumb to photon scarcity. Specifically for semantic segmentation, thermal data provides an invariant heat-signature prior very helpful for class-specific identification (e.g., distinguishing biological entities from inanimate structures). 

\textbf{Improving Segmentation Performance Through Enhancement. } VesperMUSE adopts a pre-segmentation thermal-guided enhancement approach, where low-light images are first enhanced using a model trained separately from the semantic segmentation network. This design provides greater modularity, allowing the framework to be extended to other degradations, like haze or blur, by simply replacing the enhancement module. Another advantage is that the training data can be decoupled, removing the need for low-light, well-lit, and semantic segmentation annotations to coexist within the same dataset. Since the enhancement network is trained independently, its predictions achieve a high degree of visual quality, as the module is optimized to approximate normal lighting conditions. In contrast, jointly trained (one-stage) networks may produce intermediate-enhanced outputs represented in latent space which is not suitable for direct human evaluation. This aspect is particularly important for the present work, as the low-light enhanced RGB map is further used as input for additional representations.

\begin{figure}[t]
    \centering
    \begin{minipage}[b]{0.48\linewidth}
        \centering
        \includegraphics[width=\linewidth]{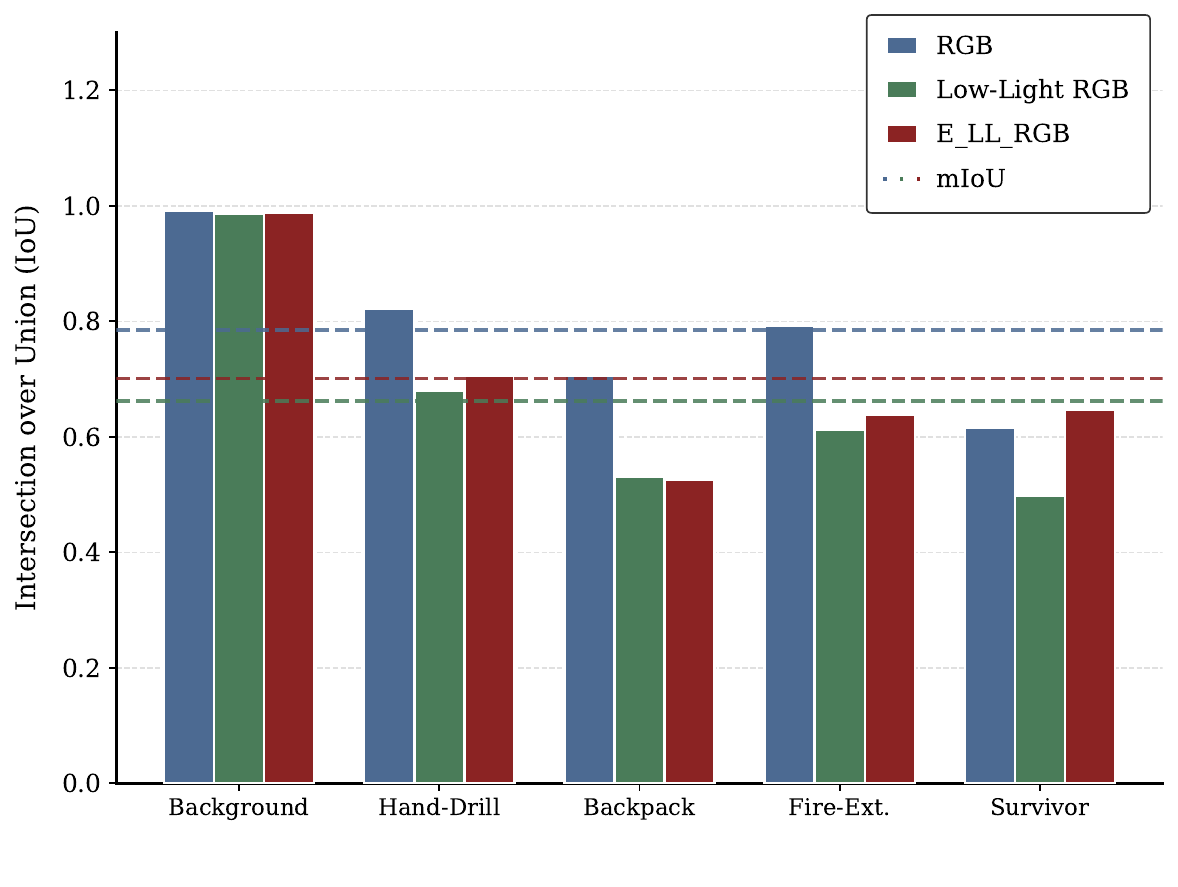}
    \end{minipage}\hfill
    \begin{minipage}[b]{0.48\linewidth}
        \centering
        \includegraphics[width=\linewidth]{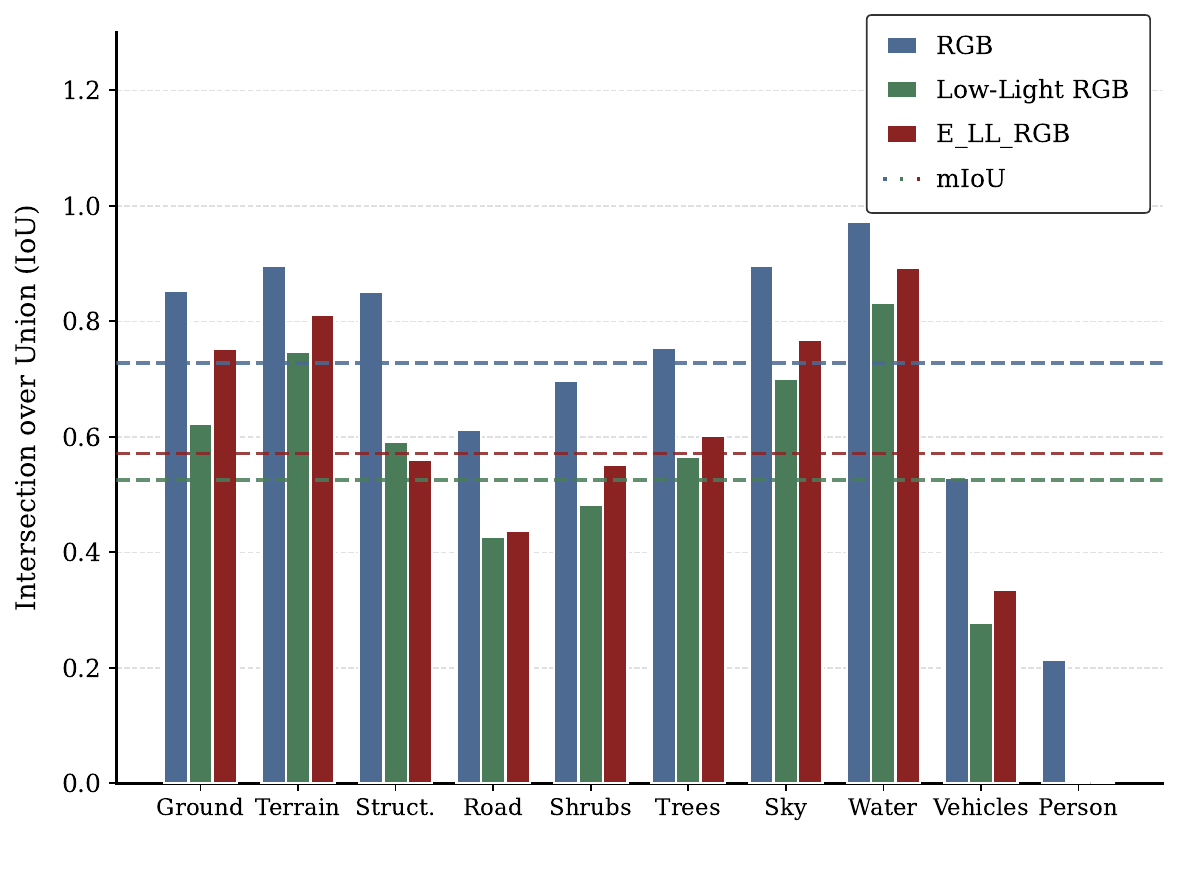}
    \end{minipage}
    \vspace{-2mm}
    \caption{Mean Intersection over Union (mIoU) benchmarking results for semantic segmentation under three input configurations, well-lit, low-light, and enhanced low-light maps, on the PST900 \cite{shivakumar2020pst900} (left) and CART \cite{lee2024cart} (right) datasets, highlighting the impact of illumination conditions and enhancement strategies on semantic segmentation performance.}
    \label{fig:intro_llie_reliability}
\end{figure}

\textbf{Challenge Understanding Through Data. } To assess the claims made regarding the issues semantic segmentation models have in low-light conditions an analysis will be conducted in Figure \ref{fig:intro_llie_reliability}  focusing on three scenarios, namely conducting segmentation on low-light data directly, LLIE imaging, and the well-lit representation respectively. For the scope of this analysis, a simple U-shaped \cite{unet} ViT \cite{dosovitskiy2020image} using different hierarchical layers of a pretrained ResNet50 \cite{koonce2021resnet} in the encoding stage and attention mechanisms on the decoding side is employed. The loss function is a hybrid combination of Dice and Focal loss assisted with the class weights derived from the training data. All three training processes are conducted with 40k iterations employing a cosine annealing learning rate scheduler with an initial $3 \times 10^{-4}$ learning rate which drops to $1 \times 10^{-7}$ by the end of training.

Observations point out that the degradation of scene illumination severely bottlenecks semantic parsing. Across both the PST900 \cite{shivakumar2020pst900} and CART \cite{lee2024cart} datasets, processing plain low-light RGB inputs results in substantial performance drops, yielding mIoU scores of only $0.6614$ and $0.5246$, respectively. This stark decline compared to the well-lit baselines underscores the necessity of the pre-segmentation enhancement strategy. By routing the degraded inputs through the independently trained enhancement module, the network successfully recovers critical scene context, elevating the segmentation performance to $0.7010$ and $0.5711$ on the two datasets. While this enhancement bridges a significant gap, it does not fully restore the optimal original well-lit RGB performance ($0.7855$ and $0.7275$).

\textbf{Further improvements. } In the context of VesperMUSE, to aid the semantic parsing process, LLIE is conducted on the low-light images with thermal being brought as an extra representation. The infrared imagery is later employed \cite{zhang2023cmx} for the segmentation training as well, since it provides relevant cues for both tasks. To make the current approach more robust, several questions need answers in order to move forward: How to integrate more structural information if thermal maps do not provide all that is needed? How will the information be fused in such a way that if specific scenarios might require more modalities, such as Light Detection and Ranging (LiDAR) \cite{raj2020survey}, or event cameras, the network will structurally support that without architecture modifications?

 The answer to the first question lies in the use of surface normal maps derived from monocular depth estimation. Specifically, DepthAnything V2 \cite{yang2024depth} is employed to extract monocular depth maps from the RGB input (which may be the original RGB or synthetic low-light adapted and enhanced depending on the configuration and objective). These monocular depth maps provide additional structural information which might be missing from the RGB and thermal maps by leveraging the pre-trained representations. Surface normals are then obtained by normalising these maps through gradient extraction.

The main innovation introduced by the core component of the VesperMUSE, its semantic segmentation module,  lies in the attention mechanism employed during the encoding stage. The underlying concept is that each modality is analyzed both globally and at the level of fine local features, thereby providing a comprehensive representation. To capture the contribution of each modality, hierarchical feature maps are fused and treated as inputs to the main attention components. Given that the attention mechanism is conceptually designed for one primary modality and $k \in \mathbb{N}$ auxiliary modalities, it can be readily configured to support an arbitrary number of modalities with varying channel configurations (e.g., RGB with three channels or single-channel thermal maps). This flexibility makes the model highly adaptable to diverse scenarios and imaging conditions where the availability of different modalities is different than the current analysed scenario, thereby addressing the second question proposed before.

\textbf{Data availability and adaptation. } One important aspect to take into account is the limited availability of data for conducting such a study. While multi-modal semantic segmentation datasets provide a high degree of reliability, and a wide range of datasets serve as effective benchmarks for LLIE, no available dataset integrates well-lit and low-light counterparts RGB, thermal imagery, structural information, and annotated semantic maps simultaneously. Given that semantic segmentation annotations are expensive to obtain, the most appropriate approach is to utilize datasets that include well-lit RGB images, thermal/structural data, and semantic maps.

Consequently, a novel synthetic low-light adaptation pipeline is proposed as part of the VesperMUSE framework. The algorithm leverages local image structure and global lighting information, generating degradations that capture natural exposure variations, noise characteristics, and colour distortions. A dedicated analysis provides a quantitative evaluation of the physical and perceptual quality of the synthesized images to show the generated images' conformity with real-world data. This integration of both local and real-world illumination information ensures that the generated images faithfully reproduce the deficiencies of natural low-light environments.
\newpage
To summarize, the contributions of VesperMUSE are:

\begin{itemize}
    \item \textbf{Unified framework:} VesperMUSE enables improved low-light semantic segmentation while preserving strong performance in well-lit settings.
    
    \item \textbf{Thermal-guided LLIE:} A highly computational-efficient enhancement module tailored to optimize downstream semantic segmentation.
    
    \item \textbf{Modality-agnostic fusion:} A scalable Fusion Vision Transformer supporting an arbitrary number of input modalities.
    
    \item \textbf{Comprehensive evaluation:} Consistent competitive performance with SOTA results across multiple benchmarks.
\end{itemize}

\chapter{Related Work}
\newpage
\section{Synthetic Low-Light Image Adaptation}
The development of LLIE models has been historically hampered by the limited dataset availability and the quality of light-deficient representations. Early efforts such as the SID dataset \cite{SID} focused on capturing short-exposure raw images along with long-exposure counterparts. This was a good representation of how a LLIE dataset should look like, however it was limited in terms of the format it used: RAW. The consumer and surveillance hardware operate much more easily using conventional formats such as PNG or JPG, which also provide more efficient computational processing. Moreover, the conversion of RAW images into other encodings fails to replicate the natural noise and illumination patterns captured in the low-light conditions, the quality of the resulting data hindering the supervised learning process of LLIE models. 

More recently, the problem with data limitation was fixed with the appearance of paired real-world datasets such as LOL \cite{wei2018deep}, LOL-v2 real \cite{yang2021sparse}, SDSD-in/out \cite{SDSD}, and LOL-Blur \cite{LOL-Blur}. Another field of concern, however, lies in the reliance of synthetic low-light adaptation methods. Widely-used current approaches, such as the work of Yang et al. through their LOL-v2 synthetic dataset \cite{yang2021sparse} have provided a very solid foundation for how synthetic low-light images should be constructed: following a real-world dataset illumination distribution extracted from multiple datasets and delivering the result into an easy-to-use encoding format.

\section{Low-Light Image Enhancement}

\noindent \textbf{Classical Methods. }
LLIE has been firstly addressed using plain-enhancement \cite{Wang_2020} which focused mainly on histogram equalisation \cite{Kansal_2018} and gamma correction  \cite{Xiao_2016, Kim_2016}. These methods amplify the brightness and contrast of underexposed images through pixel intensity transformations. Because of the complex  and uneven way illumination interacts with different areas from the image, these models struggled to model the light projection process, resulting in perceptually inconsistent outputs containing over-enhanced regions and amplified noise.

The later introduced Retinex theory \cite{Retinex} aims to decompose the image in two components: reflectance, which represents the intrinsic scene information and illuminance, focusing on lighting variations. To better account for the illumination effects on the local and global structures, LLIE-methods have employed the theory as the underlying assumption behind the image architecture \cite{RetinexNet, uretinexnet, Diff-retinex}. These models focused on estimating and refining the illumination map and reconstructing the image in a later stage with the more balanced and denoised reflectance map. This component-separation approach lead to more visually appealing and plausible results than the traditional enhancement techniques which rely heavily on hand-crafted priors and assumptions, making them hard to generalize across multiple datasets.

\noindent \textbf{CNN-based Methods. }
Besides the improvements brought in language applications or high level CV tasks, the advancement of Deep Learning (DL) has also transformed the field of image enhancement becoming the dominant paradigm \cite{Zhang_2019, Zhang_2021, Wang_2019}. One of the earliest DL models on the scene of LLIE was LLNet \cite{LLNet} which proved that supervised feature learning can improve brightness and contrast while also recovering lost details. Generative models, like EnGAN \cite{EnGAN}, focused on creating a direct mapping from low-light images to their enhanced counterparts by employing an end-to-end framework. These methods further improved upon traditional hand-crafted heuristic-based algorithms through hierarchical feature extraction.

The Retinex theory \cite{Retinex} is furthermore employed as an assumption, leveraging the ability of DL frameworks to learn feature decomposition. While such models, benefit from both physical priors and data-driven learning, they often require multi-stage training pipelines through subnetwork specialisation \cite{uretinexnet} on tasks such as: decomposition, illumination/reflectance enhancement, and reconstruction. As a result, the computational and training complexity of this kind of models are high. To solve this, one-stage Retinex-inspired Convolutional Neural Networks (CNN) have also been proposed in order to increase the computational efficiency, but they struggle with noise mitigation and artifacts, resulting in an unsatisfactory enhancement process. 

Another issue that CNN-based models exhibit is the limited receptive field of convolutional layers, hence reducing their ability to capture long-range spatial dependencies constructing the global contextual relationship across distant image regions. As light's interaction with the scene is not linear, hence why histogram-matching models fail in the LLIE domain, this limitation restricts the models' capacity of capturing diverse scene structures and mapping the relationships between them.

\noindent \textbf{ViT-based Methods. }
To overcome the inherent locality of convolutional operators, ViT-based models \cite{dosovitskiy2020image, lyt} have dominated the SOTA models charts for LLIE in the past years. The success achieved in machine translation by the Transformer architecture \cite{vaswani2017attention}, has also spread to CV and more specifically to image enhancement. The principal particularity that ViTs possess is the self-attention module which manages to capture long-range dependencies which guide the enhancement process towards a complex resolution of the LLIE challenge.

Several architectures have marked the evolution of using ViTs \cite{dosovitskiy2020image} in LLIE. For example, UFormer \cite{UFormer} adapts the U-shaped \cite{unet} encoder-decoder paradigm, replacing the standard convolutions found in U-Net \cite{unet} with Transformer blocks while preserving level-wise skip connections. Retinexformer \cite{Retinexformer} incorporates illumination representations into the self-attention mechanism to model non-local interactions across regions with varying lighting conditions. More recent work, such as ISALux \cite{balmez2026isalux} or ModalFormer \cite{brateanu2025modalformer}, successfully manage to integrate semantic and structure priors into the self-attention module aiming to compensate the light-deficiency induced detail loss with visual or feature-based information represented through different modalities.  

Despite their strong performance, Transformer-based methods face notable computational challenges. The self-attention mechanism incurs substantial computational complexity due to its reliance on the spatial dimensions of the feature maps. To mitigate this limitation, several methods, such as Restormer \cite{Restormer}, DepthLux \cite{balmez2025depthlux}, employ different strategies: computing the attention map channel wise considering the depth of the feature map as the number of channels, contracting the feature space before the self-attention module and then expanding it back to its original dimensions, or replacing convolutions with their more efficient variation, depthwise separable convolutions.

\noindent \textbf{RGB-Thermal Fusion for LLIE. }  RGB-Thermal fusion has recently been explored to enhance robustness in extreme low-light conditions by leveraging the illumination-invariant nature of thermal imagery. In particular, some studies \cite{Kumar_Panda_2025} propose a thermal-aware ViT architecture, where thermal features are explicitly embedded into the transformer pipeline to guide global feature modeling under severe illumination degradation. Complementarily, Jha et al. focus on the interaction between illumination cues derived from the low-light RGB image and the corresponding thermal map, designing a mechanism that aligns and integrates these complementary representations for more stable enhancement \cite{Jha_2025}. These works demonstrate that modeling cross-modal dependencies, especially between illumination and thermal structure, can significantly improve detail recovery and perceptual consistency in challenging low-light scenarios.

\section{Modality Fusion for Image Semantic Segmentation}
Multi-modal semantic segmentation has emerged as an effective strategy for improving scene understanding and feature richness limitations under real-world conditions where RGB might be insufficient \cite{Minaee_2021}. Complementary sensing modalities such as thermal (infrared) or depth maps have been included as inputs to improve the information representation of a specific scene \cite{zhang2023cmx}. Depth maps provide geometric and distance cues that can help disambiguate low-contrast regions and cluttered backgrounds \cite{Hu_2021}, while thermal images capture thermic distributions, offering a visual representation which is not subject to change under diverse light patterns or other deficiencies often noticed in nature like smoke or haze \cite{Kristo_2020}. Semantic segmentation is not the only high-level CV task that has been benefiting from adding multi-modality in learning, with object detection, tracking, autonomous driving perception systems \cite{Zhang_2023} also improving performance by leveraging multiple representations of the same scenes.

In the case of RGB-Depth semantic segmentation, early methods focused on fusing modalities at the input or feature level with some approaches directly concatenating RGB and depth maps or fusing them by summation \cite{Cao_2021, hazirbas2016fusenet}. In time, models have evolved to a two-stream architecture \cite{ha2017mfnet}, where independent networks extract features separately before integrating them in the decoding process. The recent focus has been placed on feature fusion approach which separates the fusion logic from the backbone and employing modules like Separation-and-Aggregation Gate \cite{Chen_2020} or Attention Complementary Modules \cite{Hu_2019} to provide parallel modality streams at each hierarchical level.

A significant focal point of recent work is discriminative treatment of modalities \cite{Zhao_2024}. This represents a change from the initial models which treated all network layers equally. More recent frameworks like LASNet \cite{chen2022lasnet} argue for level-specific treatment. This involves designing distinct models for different hierarchies: enhancing shallow spatial details (thermal-focused) and deep semantic features (RGB-focused) separately. Methods frequently integrate channel weighted selection and spatial-attention mechanisms to mitigate the inherent discrepancies (originating from distinct underlying data distributions) between modalities \cite{Zhang_2021}.  

The inherent limitations of CNNs, stemming from their constrained receptive fields, are also evident in modality fusion models. The Transformer architecture \cite{dosovitskiy2020image} has emerged as a compelling solution to this issue \cite{Zheng_2024}. Driven by the capacity of the self-attention mechanism to model complex relationships across different contexts, current research is shifting from dedicated dual-modality models toward unified architectures. While earlier methods perform well on specific pairings like RGB-Depth or RGB-Thermal \cite{hu2019acnet}, unified frameworks capable of integrating multiple concurrent data streams offer a much richer understanding of scene context, semantics, and structure. Consequently, these architectures yield superior performance on advanced scene understanding tasks. 

\chapter{Methods}
\newpage
\section{Framework Design and Motivation}

The framework behind VesperMUSE is based on three main components each having a dedicated role in the focus of this study: finding solutions for improving semantic segmentation under both low-light and well illuminated scenes.

VesperMUSE can be run both in low-light or normal light configuration. If the latter is activated, then the framework starts from Stage $3$.

\textbf{Stage 1: Low-Light Synthetic Adaptation. } The first step is obtaining the synthetic low-light imaging which will be processed using the algorithm presented in Section \ref{sec:synthetic_ll_algo}. To link this framework as closely as possible to real world scenarios, the reasoning was physics-informed. 

    \textbf{Stage 2: Low-Light Image Enhancement. } After the low-light maps are processed they serve as inputs to the thermal-guided LLIE module described in Section \ref{sec:llie_module}, along with the thermal imaging. The LLIE model is trained under a supervised approach trying to predict the original RGB map coming from the two focus datasets: PST900 \cite{shivakumar2020pst900} and CART \cite{lee2024cart}. The training and test splits are based on the ones used for the semantic segmentation stage. The result of the training process is the best checkpoint in terms of Peak-Signal-to-Noise-Ratio (PSNR), each dataset having its own selected checkpoint.

\textbf{Stage 3: Obtaining the Surface Normal maps. } The RGB image (LLIE or not) serves as the input to the DepthAnything V2 \cite{yang2024depth} model, the resulting map being a monocular depth estimation. This representation is then used to produce surface normals by taking the gradients of it and thereby aiming to normalise the structural imagery.

\textbf{Stage 4: Fusion-based Semantic Segmentation. } The three inputs corresponding to the same scene: RGB, thermal, and surface normal are then passed to the Fusion Vision Transformer which processes each of them using a dedicated pre-trained backbone and then fuses them in a modality-agnostic manner. After the features have been fused, a decoding process commences processing the output of the framework, the segmented map. Likewise LLIE, each dataset has its own trained checkpoint selected on the mIoU performance.

\section{Modalities}
\subsection{Synthetic low-light adaptation algorithm}
\label{sec:synthetic_ll_algo}
 To show that semantic segmentation on low-light images can be improved using additional modalities, it required a robust real-environments informed synthesis pipeline. The current methods are limited in terms of: input data format (relying on the RAW format), noise modeling (linear noise used as implicit), global histogram matching (introducing unwanted artifacts, as colour and luminance information are not separated), and fixed parameter configuration (controlling environment variables is impossible).

As a result, VesperMUSE framework proposes a novel pipeline for obtaining synthetic low-light images from RGB data. The algorithm aims to achieve a physics-informed degradation, accounting for the effect low-light conditions have on the visual representations. 

\noindent \textbf{Real-World Illumination Distribution. } First of all, based on a significantly broader range of real-world datasets compared to prior work, namely (LOL \cite{wei2018deep}, LOL-v2 real \cite{yang2021sparse}, SDSD-in/out \cite{SDSD}, LOL Blur \cite{LOL-Blur}), a cumulative illumination distribution is computed using the Y channel obtained after conversion to the YCbCr colour space. This choice is motivated by the fact that YCbCr provides a natural representation of illumination through its luminance component. 

Let $\Gamma$ denote the transformation from RGB to YCbCr, and let $\Lambda_Y$ represent the projection operator that extracts the luminance (Y) channel from the transformed representation. Considering the operator $\Gamma_Y$ that converts an RGB image $\mathcal{I} \in \mathbb{R}^{H \times W \times 3}$, with $H$ and $W$ denoting the image height and width, respectively, and separates the Y channel, the luminance extraction can be formally defined:

\begin{equation}
\Gamma_Y : \mathbb{R}^{H \times W \times 3} \rightarrow \mathbb{R}^{H \times W}, 
\quad
\Gamma_Y(\mathcal{I}) = \Lambda_Y\big(\Gamma(\mathcal{I})\big).
\end{equation}

Using the Y dimensionality of samples, the empirical cumulative distribution function (CDF) of luminance over the entire multi-dataset collection can be defined as follows. Let $K$ denote the total number of images in the collection, $\Omega_k$ the set of pixel coordinates in image $\mathcal{I}_k$, and $N = \sum_{k=1}^{K} |\Omega_k|$ the total number of pixels across all images. Then, the CDF that represents the cumulative proportion of pixels in the dataset is defined as:

\begin{equation}
\mathcal{F}_Y = \frac{1}{N} \sum_{k=1}^{K} \Big| \big\{ x \in \Omega_k \big\} \Big|.
\end{equation}

After the distribution over the low-light image has been computed, each original RGB image from the 2 datasets used: PST900 \cite{shivakumar2020pst900} and CART \cite{lee2024cart}, is passed through the following pipeline. 

\noindent \textbf{Retinex Decomposition}. To better approximate illumination variations under low-light conditions, a Single-Scale Retinex (SSR) \cite{hines2005single} approximation is employed to decouple each image into illumination and reflectance components. Given the luminance channel $Y$ of an input image $\mathcal{I} \in \mathbb{R}^{H \times W \times 3}$, the decomposition assumes that the observed intensity at pixel $x$ can be expressed as the product of illumination $L(x)$ and reflectance $R(x)$:

\begin{equation}
Y(x) = L(x) \cdot R(x).
\end{equation}

An important point about the luminance channel $Y$ is that even though it encodes brightness, it is composed of illumination (lighting) and reflectance elements (texture, albedo) and it is not equivalent with the illumination component assumed in the Retinex Theory \cite{Retinex}.

For the purpose of this study, the illumination component is approximated using a Gaussian smoothing operator ($\mathcal{G}_{\sigma}$) with kernel standard deviation $\sigma = 15$, reflecting the assumption that illumination varies slowly across the spatial domain. The reflectance component, which captures high-frequency structural details, is obtained via point-wise division, where $\epsilon$ is a small constant introduced for numerical stability:

\begin{align}
L &= \mathcal{G}_{\sigma}(Y), \quad
R = \frac{Y}{L + \epsilon}.
\end{align}

\begin{figure}[t]
    \centering
    \includegraphics[width=0.7\linewidth]{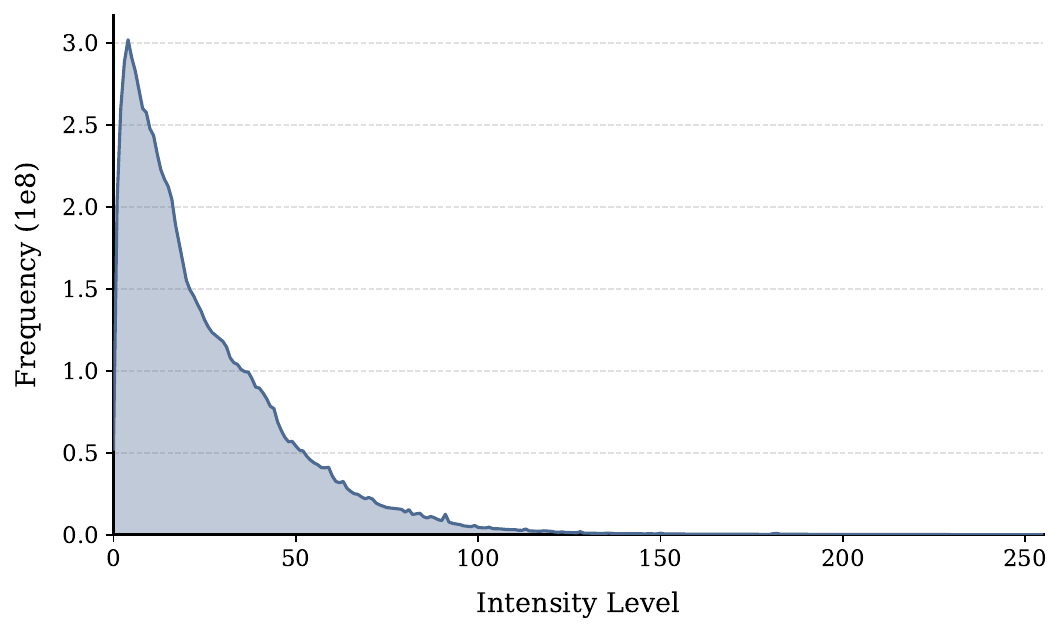}
    \vspace{-2mm} 
    \caption{An illustration of the real-world light distribution which is followed in the Real-world Distribution Matching stage}
    \label{fig:real_world_light_distribution}
\end{figure}

\noindent \textbf{Real-world Distribution Matching.} 
After decomposing the image into illumination and reflectance components, the illumination channel can be manipulated independently while preserving structural details. To emulate real-world low-light conditions (Figure \ref{fig:real_world_light_distribution}), the illumination map $L(x)$ is adjusted to match the cumulative luminance distribution $\mathcal{F}_Y$. This is achieved using histogram specification:

\begin{equation}
L_{\text{matched}}(x) = \mathcal{F}_Y^{-1}\Big( \mathcal{F}_L\big( L(x) \big) \Big).
\end{equation}

\noindent \textbf{Synthetic Luminance Reconstruction.} 
Once the illumination has been adjusted, the synthetic low-light luminance channel is reconstructed by combining the matched illumination with the original reflectance $R(x)$, ensuring that the image retains its original structural details.

\begin{equation}
Y_{\text{retinex}}(x) = L_{\text{matched}}(x) \cdot R(x).
\end{equation}

Because histogram matching can introduce nonlinear intensity remapping and alter the global energy of the illumination component, it may produce luminance values that deviate from physically plausible exposure changes and lead to inconsistencies in overall brightness. There is therefore a need to enforce a globally consistent exposure adjustment that compensates for shifts in illumination statistics while preserving structural fidelity. A mean-based scaling factor is derived from the ratio between the matched and original illumination components, ensuring that the synthetic luminance remains photometrically coherent and consistent with realistic camera exposure behavior.

\begin{equation}
s = 
\frac{\mathbb{E}\!\left[L_{\text{matched}}\right]}
     {\mathbb{E}\!\left[L\right] + \varepsilon}.
\end{equation}

\begin{equation}
Y_{\text{linear}}(x) = s \cdot Y(x).
\end{equation}

\noindent \textbf{Structure-Aware Luminance Blending.} The main advantage of the proposed method is taking into account the current luminance level of images. If the conditions are already light-deficient due to real-environment conditions, the synthetic low-light degradation can be tuned to be low-impact.

A spatially adaptive blending mask is introduced to reconcile the complementary properties of the two luminance reconstructions. While the Retinex-based luminance enforces realistic illumination statistics, its spatially varying remapping may compress bright structures and attenuate salient edges. In contrast, the globally scaled luminance preserves structural consistency but does not fully reflect real-world low-light illumination distributions. To prevent over-suppression of high-intensity regions and to maintain perceptual stability in structurally significant areas, a highlight-aware blending mechanism is employed.

Specifically, let $\tau = 150$ denote the luminance threshold beyond which highlight preservation becomes increasingly important, and let $\kappa = 100$ control the smooth transition range controlling how rapidly the blending weight increases from 0 to 1. Incorporating a structure bias parameter $\beta \in [0,1]$ guarantees a minimum contribution from the globally scaled branch. The mask is used to balance between the two illumination components:

\begin{equation}
M(x) =
\max\!\left(
\operatorname{clip}\!\left(
\frac{Y(x) - \tau}{\kappa},
0, 1
\right),
\beta
\right).
\end{equation}

\begin{equation}
Y_{\text{final}}(x)
=
M(x)\,Y_{\text{linear}}(x)
+
\big(1 - M(x)\big)\,Y_{\text{retinex}}(x).
\end{equation}

\noindent \textbf{Colour Re-integration.}
Before applying global exposure adjustments and simulating sensor noise, the synthetic luminance channel must be recombined with the original chrominance information. The synthetic low-light RGB image, $\mathcal{I}_{\text{low\_rgb}} \in \mathbb{R}^{H \times W \times 3}$, is reconstructed by merging $Y_{\text{final}}$ with the original $Cr$ and $Cb$ channels and applying the inverse colour space transformation $\Gamma^{-1}$:

\begin{equation}
\mathcal{I}_{\text{low\_rgb}}(x) = \Gamma^{-1}\Big( Y_{\text{final}}(x), Cr(x), Cb(x) \Big).
\end{equation}

\noindent \textbf{Exposure Scaling (optional). }
Given the capturing conditions of real-world low-light images, which might include constant settings or scenarios, the pipeline includes a random sampling step. A global dimming factor is applied directly to the 3-channel RGB image to add variability and diversity in exposure patterns. Given the exposure range $[s_{\min}, s_{\max}]$, the scaled image is computed as:

\begin{equation}
\mathcal{I}_{\text{scaled}}(x) = \operatorname{clip}\Big( s_r \cdot \mathcal{I}_{\text{low\_rgb}}(x), \; 0, 255 \Big), \quad s_r \sim \mathcal{U}(s_{\min}, s_{\max}).
\end{equation}

\noindent \textbf{Noise Simulation. } 
In low-light conditions, image sensors exhibit two primary types of noise: photon shot noise, caused by the discrete nature of light, and read noise, introduced by the sensor electronics. To realistically model these effects, the gamma-compressed RGB image $\mathcal{I}_{\text{scaled}}(x)$ (which is encoded in the 0--255 range) is first converted into a physically meaningful linear intensity representation $\mathcal{I}_{\text{linear}}(x) \in \mathbb{R}^{H \times W \times 3}$. 

This linearisation ensures that the added noise correctly reflects the underlying photon and electronic statistics across all three colour channels, naturally simulating real-world chroma noise. The conversion is performed by normalizing $\mathcal{I}_{\text{scaled}}$ to $[0,1]$ and inverting the gamma with an exponent of $2.2$, a common approximation for sRGB images:

\begin{equation}
\mathcal{I}_{\text{linear}}(x, c) = \left(\frac{\mathcal{I}_{\text{scaled}}(x, c)}{255}\right)^{2.2}, \quad c \in \{R, G, B\}.
\end{equation}

Noise is then added to this linear signal independently for each channel. Photon shot noise is modeled as a Poisson process scaled by $\lambda = 300$, and read noise is modeled as zero-mean Gaussian noise with standard deviation $\sigma = 0.01$. A tunable gain factor $g$ controls the overall noise intensity:

\begin{equation}
\mathcal{I}_{\text{noisy}}(x, c) = \mathcal{I}_{\text{linear}}(x, c) 
                       + g \cdot \Bigg( 
                         \frac{\text{Poisson}\big(\mathcal{I}_{\text{linear}}(x, c) \cdot \lambda\big)}{\lambda} 
                         - \mathcal{I}_{\text{linear}}(x, c) 
                         + \mathcal{N}(0, \sigma^2) 
                         \Bigg).
\end{equation}

The noisy linear signal is then clipped to $[0,1]$ and transformed back to the sRGB domain using standard gamma correction:

\begin{equation}
\mathcal{I}_{\text{srgb}}(x, c) = 255 \cdot \Big( \operatorname{clip}(\mathcal{I}_{\text{noisy}}(x, c), 0, 1) \Big)^{\frac{1}{2.2}}.
\end{equation}

\noindent \textbf{Hot Pixel Removal.}
Because statistical noise simulation can occasionally produce isolated, extreme-intensity pixels (hot pixels) atypical for modern sensors, a conditional median filter is applied as a final post-processing step. Let $\mathcal{M}_{3 \times 3}$ denote a $3 \times 3$ median filter. To preserve high-frequency structural details across the majority of the image, the filter is only applied to replace pixels exceeding a high-intensity threshold ($> 250$):

\begin{equation}
\mathcal{I}_{\text{final}}(x, c) =
\begin{cases}
\mathcal{M}_{3 \times 3}\big(\mathcal{I}_{\text{srgb}}\big)(x, c), & \text{if } \mathcal{I}_{\text{srgb}}(x, c) > 250. \\
\mathcal{I}_{\text{srgb}}(x, c), & \text{otherwise}.
\end{cases}
\end{equation}

\subsection {Auxiliary modalities}

VesperMUSE's segmentation module works with three modalities: the main modality (well-lit RGB or LLIE map obtained with the algorithm presented in Section \ref{sec:synthetic_ll_algo} and the thermal-guided enhancement model described in Section \ref{sec:llie_module}), and two auxiliary modalities: thermal and surface normal derived from the monocular depth map obtained from the main RGB modality.

\noindent \textbf{Thermal. } As previously discussed, the thermal modality provides enhanced robustness under varying illumination conditions. Furthermore, its effectiveness is largely attributed to the intrinsic physical properties it captures, namely the thermal inertia and emissivity of different materials and surfaces. These characteristics offer complementary information to the Fusion Vision Transformer, beyond what is available in the visible spectrum.

In the PST900 dataset \cite{shivakumar2020pst900}, which depicts a subterranean environment with minimal heat sources, thermal imaging acts as a reliable structural cue, enabling the detection of humans even in the presence of obscurants. Conversely, in the CART dataset \cite{lee2024cart}, the thermal modality facilitates the discrimination of diverse textures commonly encountered in natural outdoor environments.

\textbf{Surface Normal. }
To better extract the context of the scene, a monocular depth map is extracted to highlight the structural and material differences between objects or surfaces. The noise and colour distortions, attenuated but still present in the enhanced RGB images and the thermal consistency of nearby objects in thermal maps might misguide the network in distinguishing edges and separation of entities. The first order gradients of the depth map are taken and  turned into unit normal vectors to isolate and emphasize the geometric discontinuities where objects might protrude from their backgrounds. Furthermore, the scale-invariant representation of the geometric features provides the robustness needed in terms of the relative position of the objects to the camera, enabling a informative features extraction for close objects, as found in the PST900 dataset \cite{shivakumar2020pst900}, and distant objects as in the CART dataset \cite{lee2024cart}.

Hence, a pretrained version of DepthAnything V2 \cite{yang2024depth}, $\mathcal{D}_\theta(\cdot)$,  which is kept frozen during training and inference is employed to extract a monocular depth map from the main RGB modality, $\mathcal{I}_{RGB} \in \mathbb{R}^{H \times W \times 3}$ 

\begin{equation}
\mathbf{D} = \mathcal{D}_\theta(\mathcal{I}_{RGB}).
\end{equation}

\vspace{0.5em}
\noindent \textbf{Surface Normal Computation.}
Minimum-maximum normalisation is performed on the monocular depth map. Given the normalised depth map $\mathbf{D}$, the surface normals are computed and normalised to $[0,1]$:
\begin{equation}
\hat{\mathbf{n}}(x,y) =
\frac{
\left(-\frac{\partial \mathbf{D}}{\partial x},\ -\frac{\partial \mathbf{D}}{\partial y},\ 1\right)
}{
\left\|\left(-\frac{\partial \mathbf{D}}{\partial x},\ -\frac{\partial \mathbf{D}}{\partial y},\ 1\right)\right\|_2
}.
\end{equation}

\section{Low-Light Image Enhancement Vision Transformer}
\label{sec:llie_module}

The architecture of the VesperMUSE LLIE module follows a U-shaped \cite{unet} encoder-decoder structure, a widely adopted design in ViT-based \cite{dosovitskiy2020image} image restoration due to its capability for hierarchical, multi-scale feature extraction and progressive feature refinement. 
The overall structure is organized into three principal components: an encoder, a bottleneck, and a decoder.

The inputs consist of a synthetic low-light RGB image, denoted as 
$\mathcal{I}_{\mathrm{RGB}} \in \mathbb{R}^{H \times W \times 3}$, 
and a corresponding thermal representation, 
$\mathcal{I}_{\mathrm{Thermal}} \in \mathbb{R}^{H \times W \times 1}$. 
Let $VesperMUSE_{\mathrm{LLIE}}$ denote the LLIE module of VesperMUSE. 
The enhancement process can be formally expressed as:

\begin{equation}
\hat{\mathcal{I}}_{\mathrm{RGB}} 
= VesperMUSE_{\mathrm{LLIE}}
\left(
\mathcal{I}_{\mathrm{RGB}}, 
\mathcal{I}_{\mathrm{Thermal}}
\right),
\quad 
\hat{\mathcal{I}}_{\mathrm{RGB}} \in \mathbb{R}^{H \times W \times 3}.
\end{equation}

As the enhancement process commences, the RGB component is first mapped into a high-dimensional latent feature space through a pointwise convolution projection, with  output depth, $C$, denoting the base embedding dimension. The projection aligns the spatial features with the transformer embedding space while preserving spatial resolution.

\begin{equation}
F^{0} = \mathrm{conv}_{1\times1}\!\left(\mathcal{I}_{\mathrm{RGB}}\right), 
\quad 
F^{0} \in \mathbb{R}^{H \times W \times C}.
\end{equation}

\noindent\textbf{Encoding Stage. } Feature extraction is performed using a two-stage hierarchical encoder composed of Thermal-aware Transformer Block described in the following section. After each block, the features are spatially reduced in terms of the height and the width by a factor of $2$ through $\mathrm{conv}_{3\times3}$ layers with a stride of $2$, but also doubled in depth, for maintaining representational power.
Each block integrates RGB features with scale-matched thermal representation. For each resolution level $k \in \{0,1\}$, the encoder generates $\mathcal{F}_{\mathrm{enc}}^{k} \in  \mathbb{R}^{\frac{H}{2^{k}} \times \frac{W}{2^{k}} \times C \cdot 2^{k}}.$

\noindent\textbf{Bottleneck. } 
At the coarsest resolution, a dedicated Thermal-aware Transformer Block processes the compressed representation to model long-range dependencies and cross-modal correlations by  aggregating global context under thermal guidance before reconstruction begins. The resulting latent tensor is yielding $\mathcal{F}_{\mathrm{latent}} \in 
\mathbb{R}^{\frac{H}{4} \times \frac{W}{4} \times 4C}.$

\begin{figure}[H]
    \centering
    \includegraphics[width=\linewidth]{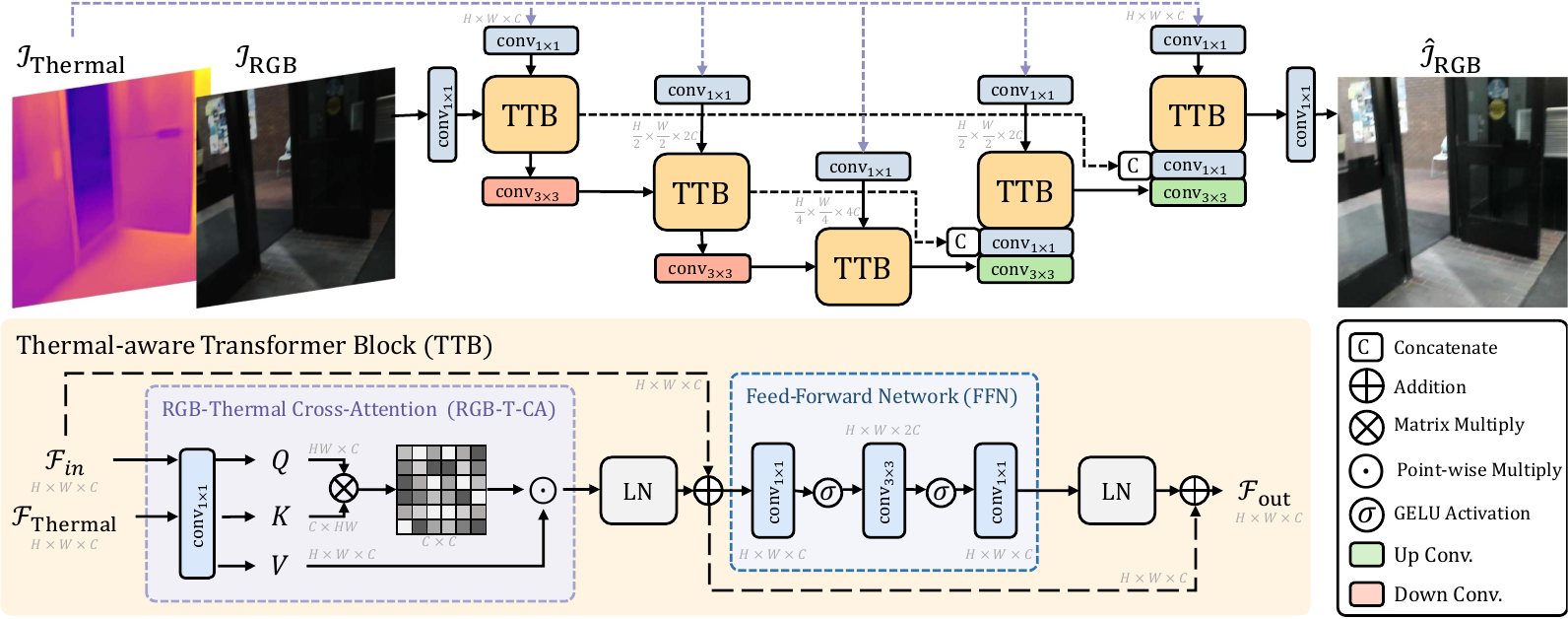}
    \vspace{-2mm} 
    \caption{Visual representation of the VesperMUSE LLIE module.}
    \label{fig:vm_llie_diagram}
\end{figure}

\noindent\textbf{Decoding stage.}
The reconstruction path mirrors the encoder through two upsampling stages. 
Each stage applies transposed $\mathrm{conv}_{3\times3}$ for spatial enlargement, concatenates the upsampled features with their encoder counterparts, and refines the fused representation using a Transformer block. For each decoding level $k \in \{1,0\}$, the representations are $\mathcal{F}_{\mathrm{dec}}^{k} 
\in 
\mathbb{R}^{\frac{H}{2^{k}} \times \frac{W}{2^{k}} \times C \cdot 2^{k}}.$

\noindent\textbf{Image Reconstruction.}
The output of the decoding phase, $\mathcal{F}_{\mathrm{final}} \in \mathbb{R}^{H \times W \times C}$, is mapped back to the RGB domain through a pointwise convolution, producing the enhanced output:

\begin{equation}
\hat{\mathcal{I}}_{\mathrm{RGB}} 
= \mathrm{conv}_{1\times1}\!\left(\mathcal{F}_{\mathrm{final}}\right),
\quad
\hat{\mathcal{I}}_{\mathrm{RGB}} 
\in 
\mathbb{R}^{H \times W \times 3}.
\end{equation}

\subsection{Thermal-aware Transformer Block}
The core component of the LLIE module is represented by the Thermal-aware Transformer Block which has 2 main constituents: the RGB-Thermal Cross-Attention (RGB-T-CA) and the Feed-Forward Network (FFN). At each step, the resulting feature maps from the two components are normalized using Layer Normalisation (LN) and added with the skip connections, originating from the previous step. Hence, with $k$ as the level of the block, $\mathcal{F}_{\mathrm{Thermal}}$ as correspondent thermal feature and $\mathcal{F}_{in}$ as the input, the functionality of the entire block can be expressed as:

\begin{gather}
\mathcal{F}'_{in} 
= 
\mathcal{F}_{in} 
+ 
\mathrm{LN}\!\left(
\mathrm{RGB\text{-}T\text{-}CA}
\left(
\mathcal{F}_{in}, 
\mathcal{F}_{\mathrm{Thermal}}
\right)
\right) ,
\\
\mathcal{F}_{out} 
= 
\mathcal{F}'_{in} 
+ 
\mathrm{LN}\!\left(
\mathrm{FFN}
\left(
\mathcal{F}'_{in}
\right)
\right),
\\
\mathcal{F}_{in}, 
\mathcal{F}_{\mathrm{Thermal}} 
\in 
\mathbb{R}^{\frac{H}{2^k} \times \frac{W}{2^k} \times C \cdot 2^k}.
\end{gather}

\subsection{RGB-Thermal Cross-Attention}
The dedicated self-attention module is the main driver of the enhancement process because it manages to capture the underlying patterns of the data which guide the restoration of lost details due to low-light conditions. Since RGB is the map which is degraded because of the direct interaction with poor-lighting, thermal (infrared) maps are employed to provide information about the local and global structure of the image leaving the colour information to be extracted from the low-light RGB map.

Given the RGB feature map 
\(
\mathcal{F}_{\mathrm{RGB}} \in \mathbb{R}^{H \times W \times C}
\)
and the thermal feature map 
\(
\mathcal{F}_{\mathrm{Thermal}} \in \mathbb{R}^{H \times W \times C},
\)
query, key, and value embeddings are obtained through pointwise convolution projection:

\begin{equation}
Q, K, V 
=
\mathrm{conv}_{1\times1}
\big(
\mathcal{F}_{\mathrm{RGB}}, 
\mathcal{F}_{\mathrm{RGB}}, 
\mathcal{F}_{\mathrm{Thermal}}
\big),
\quad
Q, K, V \in \mathbb{R}^{H \times W \times C}.
\end{equation}

The tensors are divided into $h$ attention heads, each with channel dimension $C/h$.  
Given that $\mathcal{T}_i \in \mathbb{R}^{+}$ is a learnable temperature parameter, the attention scores for each head are computed as:

\begin{equation}
\mathrm{Attn}_i 
=
\mathrm{softmax}
\left(
\mathcal{T}_i \, Q_i K_i^{\top}
\right).
\end{equation}

The attention scores are then multiplied with the thermal $V$ projection concatenated to gather the head representation back and projected using a pointwise convolution as:

\begin{equation}
\mathcal{F}_{\mathrm{Thermal\text{-}Attn}} 
=
\mathrm{conv}_{1\times1}
\Bigg(
\mathrm{Concat}\Big(\{ Attn_{i} V_i \}_{i=1}^{h}\Big)
\Bigg),
\quad
\mathcal{F}_{\mathrm{Thermal\text{-}Attn}} \in \mathbb{R}^{H \times W \times C}.
\end{equation}

\subsection{Feed Forward Network}
\label{sec:ffn}

To further refine the thermal-guided feature map, $\mathcal{F}_{\text{Thermal-Attn}} \in \mathbb{R}^{H \times W \times C}$, the FFN section consists of three consecutive convolutional layers (with $1 \times 1$ and $3 \times 3$ kernels) applied sequentially. These layers are interleaved with Gaussian Error Linear Unit (GELU) activations, denoted by $\sigma$. The FFN employed in VesperMUSE can thus be formally expressed as:

\begin{equation}
\mathcal{F}_{\mathrm{out}} = \mathrm{FFN}(\mathcal{F}_{\text{Thermal-Attn}}) =
\mathrm{conv}_{1\times1} \Big(
\sigma \big(
\mathrm{conv}_{3\times3} \big(
\sigma \big(
\mathrm{conv}_{1\times1}(\mathcal{F}_{\text{Thermal-Attn}})
\big)
\big)
\big)
\Big).
\end{equation}

\subsection{Low-Light Image Enhancement Loss function}


\noindent \textbf{Hybrid Loss. }
The LLIE objective is formulated as a weighted hybrid loss that jointly enforces pixel-level fidelity, perceptual realism, structural consistency, and semantic preservation. 
Specifically, the loss combines: 
(i) a Mean Absolute Error (MAE) reconstruction term to ensure and intensity correctness, 
(ii) a perceptual loss computed in a pretrained feature space to enhance visual quality, 
(iii) a structural similarity loss to preserve local structures, and 
(iv) a semantic consistency term to maintain high-level content alignment with the ground truth, helping for the semantic focus required by the downstream semantic segmentation module. 
The relative contribution of each component is controlled by weighting coefficients $\alpha_1, \alpha_2, \alpha_3,$ and $\alpha_4$.

\begin{equation}
\mathcal{L}_{\text{LLIE}}
=
\alpha_1 \mathcal{L}_{\ell_1}
+
\alpha_2 \mathcal{L}_{\text{perc}}
+
\alpha_3 \mathcal{L}_{\text{SSIM}}
+
\alpha_4 \mathcal{L}_{\text{sem}}.
\end{equation}


\noindent \textbf{MAE Reconstruction Loss. }
Let $\mathbf{Y} \in \mathbb{R}^{H \times W \times C}$ denote the ground-truth image, 
$\hat{\mathbf{Y}} \in \mathbb{R}^{H \times W \times C}$ the enhanced prediction, 
$N = H \times W \times C$ the total number of elements (pixels and channels), 
and $i$ the linear index over spatial and channel dimensions. 
The mean absolute error (MAE) reconstruction loss is defined as:

\begin{equation}
\mathcal{L}_{\ell_1}
=
\frac{1}{N}
\sum_{i=1}^{N}
\left|
\hat{\mathbf{Y}}_i - \mathbf{Y}_i
\right|.
\end{equation}


\noindent \textbf{Perceptual Loss. }
Let $\phi(\cdot)$ denote a fixed pretrained feature extractor (VGG-19 \cite{vgg19}, 
$\phi(\mathbf{Y}) \in \mathbb{R}^{M}$ and 
$\phi(\hat{\mathbf{Y}}) \in \mathbb{R}^{M}$ the corresponding feature representations, 
and $M$ the total number of extracted feature elements. 
The perceptual loss is defined as:

\begin{equation}
\mathcal{L}_{\text{perc}}
=
\frac{1}{M}
\left\|
\phi(\hat{\mathbf{Y}})
-
\phi(\mathbf{Y})
\right\|_2^2.
\end{equation}


\noindent \textbf{Multi-Scale SSIM Loss. }
Let $\hat{\mathbf{Y}}$ and $\mathbf{Y}$ denote the enhanced and ground-truth images, respectively, 
$J$ the total number of scales, 
$\mu_{\hat{Y}_j}$ and $\mu_{Y_j}$ the local means at scale $j$, 
$\sigma_{\hat{Y}_j}$ and $\sigma_{Y_j}$ the corresponding standard deviations, 
$\sigma_{\hat{Y}_j Y_j}$ the cross-covariance, 
$C_1, C_2, C_3$ small positive constants for numerical stability, 
and $\alpha_J, \beta_j, \gamma_j$ scale-dependent weighting exponents. 
The luminance $l_j$, contrast $c_j$, and structural $s_j$ components at scale $j$ are defined as:

\begin{equation}
l_j
=
\frac{
2 \mu_{\hat{Y}_j} \mu_{Y_j} + C_1
}{
\mu_{\hat{Y}_j}^2 + \mu_{Y_j}^2 + C_1
},
\quad
c_j
=
\frac{
2 \sigma_{\hat{Y}_j} \sigma_{Y_j} + C_2
}{
\sigma_{\hat{Y}_j}^2 + \sigma_{Y_j}^2 + C_2
},
\quad
s_j
=
\frac{
\sigma_{\hat{Y}_j Y_j} + C_3
}{
\sigma_{\hat{Y}_j} \sigma_{Y_j} + C_3
}.
\end{equation}

The Multi-Scale Structure Similarity index metric (MS-SSIM) taking into account the luminance, contrast and structure components is defined as, with the loss represented as the distance from the perfect score:

\begin{gather}
\mathrm{MS\text{-}SSIM}(\hat{\mathbf{Y}}, \mathbf{Y})
=
\left( l_J \right)^{\alpha_J}
\prod_{j=1}^{J}
\left( c_j \right)^{\beta_j}
\left( s_j \right)^{\gamma_j},
\\
\mathcal{L}_{\text{SSIM}}
=
1 - \mathrm{MS\text{-}SSIM}(\hat{\mathbf{Y}}, \mathbf{Y}).
\end{gather}


\noindent \textbf{Semantic Consistency Loss. }
While the perceptual loss enforces similarity in a generic feature space to improve visual realism, 
it does not explicitly guarantee preservation of task-relevant semantic content. 
In LLIE, an approach which produces visually appealing images through contrast and colour balance amplification, does not imply that the model is optimized to enhance images for downstream tasks. Usually, it is quite the opposite as the type of visual composition which is appealing to the human eye is not informative enough for computer scene-understanding tasks such as semantic segmentation or object detection.

To explicitly constrain the enhancement process toward semantic fidelity, 
a frozen pretrained ResNet50 \cite{koonce2021resnet} encoder $\psi(\cdot)$ is employed as a high-level feature regularizer. 
This encoder acts as a fixed semantic prior, ensuring that the enhanced image preserves discriminative object representations consistent with the ground truth. 
By aligning deep semantic embeddings, the enhancement network is guided toward solutions that not only improve visual quality but also facilitate improved optimization for subsequent segmentation models.

Let $\psi(\cdot)$ denote the frozen semantic feature extractor,
$\psi(\mathbf{Y}) \in \mathbb{R}^{K}$ and 
$\psi(\hat{\mathbf{Y}}) \in \mathbb{R}^{K}$ the corresponding high-level feature representations of the ground-truth and enhanced images, respectively, 
and $K$ the total number of semantic feature elements. 
The semantic consistency loss is defined as:

\begin{equation}
\mathcal{L}_{\text{sem}}
=
\frac{1}{K}
\left\|
\psi(\hat{\mathbf{Y}})
-
\psi(\mathbf{Y})
\right\|_1.
\end{equation}

\subsection{Low-Light Image Enhancement Training}

The LLIE module is trained for 300K iterations using the Adam optimizer \cite{kingma2014adam}
with an initial learning rate of $2 \times 10^{-4}$ and momentum parameters 
$(\beta_1, \beta_2) = (0.9, 0.999)$. Gradient clipping is applied to ensure 
stable optimization. No warm-up phase is used.

\noindent \textbf{Learning Rate Schedule.}
A two-stage cosine annealing restart schedule is employed. 
Training is divided into two consecutive cycles with a starting rate of $2 \times 10^{-4}$ increasing to $3 \times 10^{-4}$ up to $92K$ iterations, and then decreasing to $1 \times 10^{-6}$ across $208K$ iterations. During the cosine phase, the learning rate at iteration $t$ within the 
second cycle is computed as:

\begin{equation}
\eta_t = \eta_{\min} 
+ \frac{1}{2}(\eta_0 - \eta_{\min})
\left(1 + \cos\left(\frac{\pi t}{T}\right)\right).
\end{equation}

The initial constant learning rate phase enables stable large-step 
optimization and rapid convergence in the early stage. The subsequent 
cosine annealing phase progressively reduces the update magnitude, 
facilitating fine-grained parameter refinement and improved restoration fidelity.
The training is conducted on 256 × 256 patches randomly cropped from
paired low-light and normal-light images, with a batch size of 8.

\section{Fusion Vision Transformer}
\textbf{Encoding Phase.} The objective of the Fusion Vision Transformer is to effectively integrate an arbitrary number of input modalities, defined as a primary modality, $M_{\text{main}}$, and $k \in \mathbb{N}$ auxiliary modalities, $M_{\overline{1,k}}$. 

The modality-agnostic nature of the architecture is enabled by the self-attention mechanism within the Fusion Encoder Block. This design effectively integrates global and local patterns from the auxiliary modalities while remaining inherently scalable. It accommodates an arbitrary number of inputs, without requiring structural modifications to the underlying fusion method.

Features are processed using a encoder-decoder path, where the encoding process is conducted in 4 consecutive steps each of them being represented by the Fusion Encoding Block. The decoding process does not perform fusion as it process a single feature space originating from the end of the encoding stage. 

\begin{figure}[H]
    \centering
    \includegraphics[width=\linewidth]{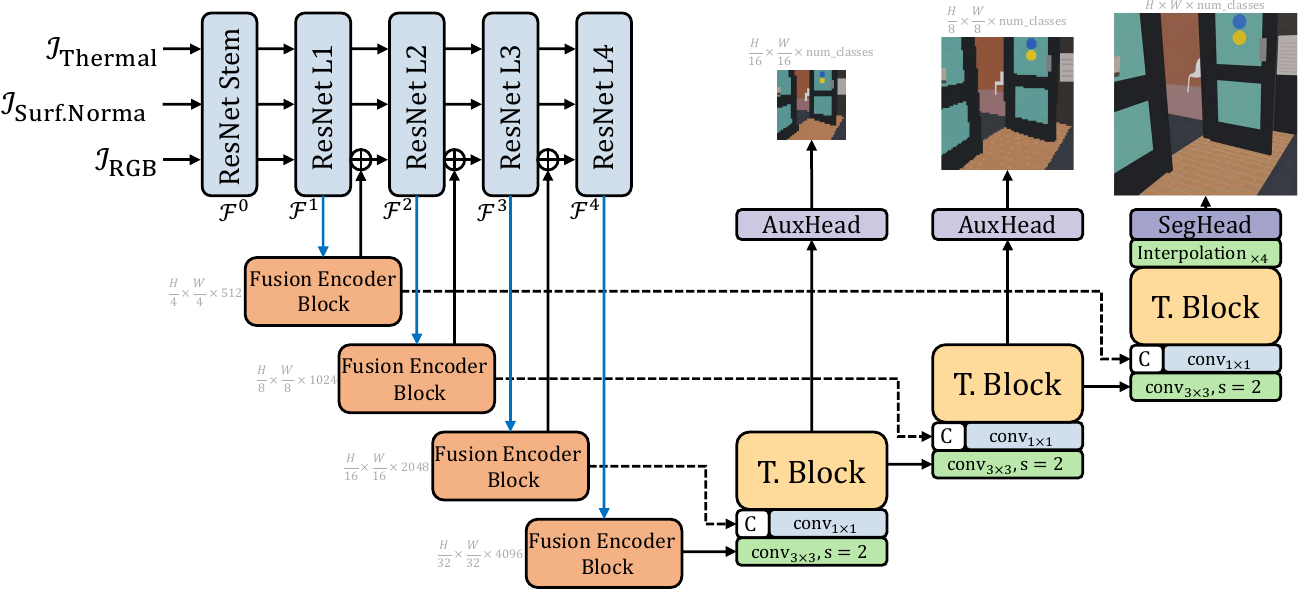}
    \vspace{-2mm} 
    \caption{Visual representation of the VesperMUSE Semantic Segmentation module.}
    \label{fig:vm_seg_diagram}
\end{figure}

During the encoding stage, the feature representations for each modality are enriched using a ResNet50 \cite{koonce2021resnet} ($\approx 25.6\text{M}$ parameters) for the primary modality and ResNet18 \cite{koonce2021resnet} ($\approx 11.7\text{M}$ parameters) backbones for the auxiliary modalities. This asymmetric backbone selection is motivated by computational efficiency, while the lighter ResNet18 \cite{koonce2021resnet} architectures minimize overhead, the deeper ResNet50 \cite{koonce2021resnet} is utilized for the RGB stream to leverage its superior representational power, which is critical for capturing the complex visual features required within the scope of VesperMUSE.

Before reaching the first Fusion Encoder Block, the inputs are formally defined as
$\mathcal{I}_{\text{RGB}} \in \mathbb{R}^{H \times W \times 3}$,
$\mathcal{I}_{\text{Thermal}} \in \mathbb{R}^{H \times W \times 1}$, and
$\mathcal{I}_{\text{Surf. Normals}} \in \mathbb{R}^{H \times W \times 3}$.
Each modality is passed through its corresponding Feature Stem Network derived from a ResNet \cite{koonce2021resnet} backbone.
Since all backbones expect three-channel inputs, the first convolutional layer in the ResNet18 \cite{koonce2021resnet} Stem, used for the thermal modality, is adapted to support single-channel inputs following a per-input-channel kernel summation approach:

\begin{equation}
\tilde{W}^{(1)} = \sum_{c=1}^{3} W^{(1)}_{c}, 
\quad \tilde{W}^{(1)} \in \mathbb{R}^{7 \times 7 \times 1 \times C}.
\end{equation}

This modification collapses the three input channels of the first convolutional layer into a single one, enabling compatibility with uni-dimensional thermal inputs while preserving the pretrained weights.

The encoding pathway of the Fusion Vision Transformer is composed of four cascaded Fusion Encoder Blocks. Each block sequentially processes the latent features through a Fusion Self-Attention module and a FFN before propagating the refined representations to the subsequent stage.

To formalize the hierarchical feature extraction across all encoding stages $k \in \{1, \dots, 4\}$, let the outputs of the initial Stem Feature Networks be denoted as:

\begin{equation}
\begin{aligned}
    &\mathcal{F}_{\text{RGB}}^{0} = \mathrm{Stem}_{\text{ResNet50}}(\mathcal{I}_{\text{RGB}}), && \mathcal{F}_{\text{RGB}}^{0} \in \mathbb{R}^{\frac{H}{4} \times \frac{W}{4} \times 64} .\\
    &\mathcal{F}_{\text{Thermal}}^{0} = \mathrm{Stem}_{\text{ResNet18}}(\mathcal{I}_{\text{Thermal}}), && \mathcal{F}_{\text{Thermal}}^{0} \in \mathbb{R}^{\frac{H}{4} \times \frac{W}{4} \times 64}. \\
    &\mathcal{F}_{\text{Surf. Normals}}^{0} = \mathrm{Stem}_{\text{ResNet18}}(\mathcal{I}_{\text{Surf. Normals}}), && \mathcal{F}_{\text{Surf. Normals}}^{0} \in \mathbb{R}^{\frac{H}{4} \times \frac{W}{4} \times 64}.
\end{aligned}
\end{equation}

The feature maps for each subsequent stage are computed by propagating the representations through the corresponding residual levels of the parallel ResNet backbones:

\begin{equation}
\begin{aligned}
    &\mathcal{F}_{\text{RGB}}^{k} = \mathrm{ResNet\_Layer}_{50}^{k}\left(\mathcal{F}_{\text{RGB}}^{k-1}\right), && \mathcal{F}_{\text{RGB}}^{k} \in \mathbb{R}^{\frac{H}{2^{k+1}} \times \frac{W}{2^{k+1}} \times (256 \cdot 2^{k-1})}. \\
    &\mathcal{F}_{\text{Thermal}}^{k} = \mathrm{ResNet\_Layer}_{18}^{k}\left(\mathcal{F}_{\text{Thermal}}^{k-1}\right), && \mathcal{F}_{\text{Thermal}}^{k} \in \mathbb{R}^{\frac{H}{2^{k+1}} \times \frac{W}{2^{k+1}} \times (64 \cdot 2^{k-1})}. \\
    &\mathcal{F}_{\text{Surf. Normals}}^{k} = \mathrm{ResNet\_Layer}_{18}^{k}\left(\mathcal{F}_{\text{Surf. Normals}}^{k-1}\right), && \mathcal{F}_{\text{Surf. Normals}}^{k} \in \mathbb{R}^{\frac{H}{2^{k+1}} \times \frac{W}{2^{k+1}} \times (64 \cdot 2^{k-1})}.
\end{aligned}
\end{equation}

Furthermore, after the block-correspondent features $\mathcal{F}_{\text{RGB}}^{k}, \mathcal{F}_{\text{Thermal}}^{k}, \mathcal{F}_{\text{Depth}}^{k}$ have been obtained, the thermal and surface normals modalities are projected using $\mathrm{conv}_{1\times1}$ in order to match the expected richer dimensionality of ResNet50 \cite{koonce2021resnet} (as the auxiliary modalities are passed through ResNet18 \cite{koonce2021resnet} modules which have different channel sizes at equivalent hierarchical levels). 

The purpose of this projection is not solely addressing shape mismatches, but also allowing the model to learn the shared latent space between all the modalities, and adapting the modalities to the expected patterns expected by the backbones, as the nature of the data distribution is different in the context of working different modalities.

\begin{equation}
\begin{aligned}
    &\hat{\mathcal{F}}_{\text{Thermal}}^{k} = \textit{proj}_{T}^{k}\left(\mathcal{F}_{\text{Thermal}}^{k}\right), && \hat{\mathcal{F}}_{\text{Thermal}}^{k} \in \mathbb{R}^{\frac{H}{2^{k+1}} \times \frac{W}{2^{k+1}} \times (256 \cdot 2^{k-1})}. \\
    &\hat{\mathcal{F}}_{\text{Surf. Normals}}^{k} = \textit{proj}_{SN}^{k}\left(\mathcal{F}_{\text{Surf. Normals}}^{k}\right), && \hat{\mathcal{F}}_{\text{Surf. Normals}}^{k} \in \mathbb{R}^{\frac{H}{2^{k+1}} \times \frac{W}{2^{k+1}} \times (256 \cdot 2^{k-1})}.
\end{aligned}
\end{equation}

Next, the extracted features enter the Fusion Encoder Block. This component is treated as a black box for now, with its detailed architecture explained in subsequent sections.

\begin{equation}
\mathcal{F}_{\text{Fused}}^{k} 
= 
\mathrm{FusionEnc}_{k}
\left(
\mathcal{F}_{\text{RGB}}^{k}, 
\hat{\mathcal{F}}_{\text{Thermal}}^{k}, 
\hat{\mathcal{F}}_{\text{Surf. Normals}}^{k}
\right).
\end{equation}

After the fused map has been computed and before continuing to the subsequent Fusion Encoder Block, the RGB future map $\mathcal{F}_{\text{RGB}}^{k}$ is updated using $\mathcal{F}_{\text{Fused}}^{k}$, transmitting the current RGB information along with the obtained fusion map:

\begin{equation}
\mathcal{F}_{\text{RGB}}^{k+1} = \mathcal{F}_{\text{Fused}}^{k} + \mathcal{F}_{\text{RGB}}^{k}.
\end{equation}

\noindent \textbf{Decoding Phase.} 
The fusion mechanism of the proposed Fusion Vision Transformer is performed during the encoding stage, while the decoding stage is dedicated to feature refinement and coherence enhancement. The decoding process is structured around a series of blocks that progressively refine the representations across three hierarchical spatial levels. At each level, skip connections from the corresponding encoding stages are concatenated with the outputs of the decoding blocks. This design facilitates stable gradient propagation and mitigates the information bottleneck that typically arises when integrating heterogeneous visual modalities characterized by distinct underlying probability distributions.

Considering the encoding stage ended, $\mathcal{F}_{\text{bottleneck}}$, the output of the last Fusion Encoder Block, is upsampled,  concatenated with the correspondent skip connection from the encoding stage, and passed through the Decoder Block (Section \ref{sec:decoder}). This process repeats three times. Given that $D$ is the depth used to project the fused map for the Segmentation Head (Section \ref{sec:seg_head}), the decoding process output can be represented as:

\begin{equation}
\begin{aligned}
    &\mathcal{F}_{\text{Decoded}} = \mathrm{Decoder}(\mathcal{F}_{\text{Bottleneck}}), && \mathcal{F}_{\text{Decoded}} \in \mathbb{R}^{\frac{H}{4} \times \frac{W}{4} \times D}.
\end{aligned}
\end{equation}

The obtained feature map, $\mathcal{F}_{\text{Decoded}}$, is then upsampled using interpolation with a factor of 4 to obtain the final fusion feature map.

\begin{equation}
\begin{aligned}
    &\mathcal{F}_{\text{out}} = \operatorname{Interpolation}_{\times 4}(\mathcal{F}_{\text{Decoded}}), && \mathcal{F}_{\text{out}} \in \mathbb{R}^{H \times W \times D}.
\end{aligned}
\end{equation}

\subsection {Fusion Encoder Block}
The Fusion Encoder Block represents the unitary component of the encoding stage of the Fusion Vision Transformer. Its role is to coherently combine the features originating from multiple modalities and to assess the unique information each one of them contains and how this information interacts with the other sets of specific data from the other modalities. 

To achieve this, the Fusion Encoder Block employs 2 modules: the Fusion Self-Attention Module which combines and selects the most relevant features based on the current context and a FFN which further refines the fused maps. The modules interact through skip connection which aid the gradient flow and normalisation layers that provide data distribution unity across the samples. 

The principal advantage of the proposed Fusion Encoder Block lies in its structural scalability to an arbitrary number of modalities, $k \in \mathbb{N}$. The only requirement is the designation of one modality as the primary modality, while the remaining $k$ modalities are treated as auxiliary. Consequently, the functionality of the Fusion Encoder Block can be formally generalized at an abstract level to accommodate any number of modalities, and can be expressed as follows:

\begin{align}
\mathcal{F}_{\text{Fused}} 
&= \mathcal{F}_{\text{RGB}} 
+ \mathrm{LN}\!\left(
\mathrm{Fusion\_SA}\!\left(\left\{ \mathcal{F}_{m} \right\}_{m=1}^{k} \right)
\right), \\
\mathcal{F}_{\text{out}} 
&= \mathcal{F}_{\text{Fused}} 
+ \mathrm{LN}\!\left(
\mathrm{FFN}\!\left(\mathcal{F}_{\text{Fused}}\right)
\right).
\end{align}

\subsection{Fusion Self-Attention Module}
The main constituent of Vision Transformers is the self-attention module which captures relevant features from the images, highlighting the prevalent information from the visual space. However, its effectiveness represents a compromise in terms of computational complexity with the standard vision self attention yielding a complexity of $\mathcal{O}(H^2 W^2)$, where $H$ and $W$ represent the height and the width of the image. To partially mitigate this problem, the Fusion Self-Attention Module adheres to the solution proposed by Restormer \cite{Restormer}, where the attention scores produce a feature map in $\mathbb{R}^{C \times C}$, hence the complexity of the the self-attention module becoming $\mathcal{O}(HWC)$, where $C$ represents the projection size of the Query ($Q$), Key ($K$), and Value ($V$) components. 

Besides computational complexity, most self-attention modules fail to integrate both local and global features from the image, treating the projection of the images as a unitary feature. The accuracy of the down-stream task on which VesperMUSE focuses on, image semantic segmentation, heavily relies on local components such as edges, isolated illumination patterns, and noise. The proposed Fusion Self-Attention Module aims to achieve a both global and local feature extraction, which allows the model to adapt the latent representations considering a wider range of information.

\vspace{3mm}
\begin{figure}[H]
    \centering
    \includegraphics[width=\linewidth]{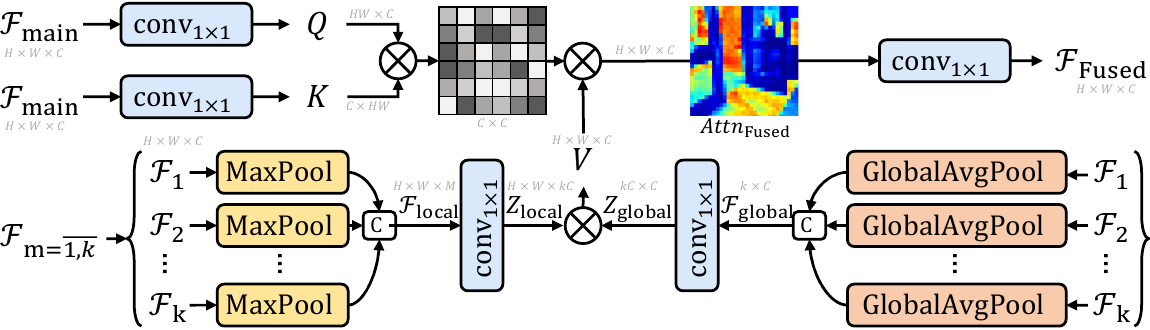}
    \vspace{-2mm} 
    \caption{Visual representation of the VesperMUSE modality-agnostic attention mechanism.}
    \label{fig:vm_llie_diagram}
\end{figure}

Hence, considering $\mathcal{F}_{main} \in \mathbb{R}^{H \times W \times C}$ as the primary modality and $\{ {F}_m \;|\; {F}_m \in \mathbb{R}^{H \times W \times C}, \; m = \overline{1,k} \}$ the set of auxiliary modalities, the general formulation of the Fusion Self-Attention Module is:

\begin{equation}
    \mathcal{F}_{\text{Fused}} = \mathrm{Fusion\_SA}\Big(\mathcal{F}_{\text{main}}, \; \{ \mathcal{F}_m \;|\; \mathcal{F}_m \in \mathbb{R}^{H \times W \times C}, \; m = \overline{1,k} \} \Big).
\end{equation}

The main modality, $\mathcal{F}_\text{main}$, serves as the source for both the Q and K representations which are obtained by projecting the main feature using two pointwise convolutional layers.

\begin{equation}
\begin{aligned}
Q &= \mathrm{conv}_{1\times1}\!\left(\mathcal{F}_{\text{main}}\right), \\
K &= \mathrm{conv}_{1\times1}\!\left(\mathcal{F}_{\text{main}}\right).
\end{aligned}
\end{equation}

The V component is obtained using both local and global information from all the auxiliary modalities. This approach facilitates better feature capturing and interaction.  The local feature extraction for the auxiliary modalities collection is done as follows:

\begin{equation}
\left\{
\max_{c} \, {F}_m 
\;\middle|\;
{F}_m \in \mathbb{R}^{H \times W \times C},
\; m = \overline{1,k}
\right\}
\;\longrightarrow\;
\left\{
\hat{\mathcal{F}}_m 
\;\middle|\;
\hat{\mathcal{F}}_m \in \mathbb{R}^{H \times W \times 1},
\; m = \overline{1,k}
\right\}.
\end{equation}

Since the individual local representations are not useful alone in a fusion context, they are concatenated to obtain a unitary map which is then projected into a higher feature space.

\begin{equation}
\begin{aligned}
    &\mathcal{F}_{\text{local}} = \mathrm{Concat}_{m=1}^{k}\left(\hat{\mathcal{F}}_m\right), && \mathcal{F}_{\text{local}} \in \mathbb{R}^{H \times W \times M}, \\
    &Z_{\text{local}} = \mathrm{conv}_{1 \times 1}\left(\mathcal{F}_{\text{local}}\right), && Z_{\text{local}} \in \mathbb{R}^{H \times W \times kC}.
\end{aligned}
\end{equation}

Following a similar approach, but with a different extraction technique, the global features are extracted using global pooling on the modality features.

\begin{equation}
\left\{ \frac{1}{HW} \sum_{h=1}^{H} \sum_{w=1}^{W} \mathcal{F}_m \;\middle|\; \mathcal{F}_m \in \mathbb{R}^{H \times W \times C}, \; m = \overline{1,k} \right\} 
\;\longrightarrow\; 
\left\{
\hat{\mathcal{F}}_m 
\;\middle|\;
\hat{\mathcal{F}}_m \in \mathbb{R}^{C},
\; m = \overline{1,k}
\right\}.
\end{equation}

The global collection of features are concatenated as well and projected to a higher dimensionality:

\begin{equation}
\begin{aligned}
    &\mathcal{F}_{\text{global}} = \mathrm{Concat}_{m=1}^{k}\left(\hat{\mathcal{F}}_m\right), && \mathcal{F}_{\text{global}} \in \mathbb{R}^{k \times C}, \\
    &Z_{\text{global}} = \mathrm{conv}_{1 \times 1}\left(\mathcal{F}_{\text{global}}\right), && Z_{\text{global}} \in \mathbb{R}^{kC \times C}.
\end{aligned}
\end{equation}

In order to combine the local and global features, the two projections are multiplied through matrix multiplication to obtain a contiguous auxiliary modality map which will represent the V component of the Fusion Self-Attention module:

\begin{equation}
V = Z_{\text{local}} \; \cdot \; Z_{\text{global}}, 
\qquad V \in \mathbb{R}^{H \times W \times C}.
\end{equation}

To reduce the computational complexity and allow better contextual interaction, the three components Q, K, V are split into $h$ distinct heads.

\begin{equation}
Q, K, V \;\longrightarrow\; \{ Q_i, K_i, V_i \}_{i=1}^{h}, 
\quad Q_i, K_i, V_i \in \mathbb{R}^{H \times W \times \frac{C}{h}}.
\end{equation}

Given that the components are computed and split, and $\mathcal{T}$ as the temperature which controls how the attention scores are balanced, the self-attention formulation is defined as:

\begin{equation}
\mathrm{Attn}_{\text{Fused}} = \mathrm{softmax}\!\Bigg(\frac{Q \cdot K^\top}{\mathcal{T}}\Bigg) \cdot V,
\quad
Q, K, V \in \mathbb{R}^{H \times W \times \frac{C}{h}}, \; \mathcal{T} \in \mathbb{R}^+.
\end{equation}

To finalise the Fusion Self-Attention Module the final map is projected back to the original depth dimensionality using a pointwise convolution.

\begin{align}
\mathcal{F}_{Fused} &= \mathrm{conv}_{1\times1}\!\left(Attn_{\text{Fused}}\right), &
Attn_{\text{Fused}} &\,\in\, \mathbb{R}^{H \times W \times C}.
\end{align}

\subsection{Feed Forward Network} The FFN used in the Fusion Vision Transformer is the same used in the LLIE module, with the formal description found in Section \ref{sec:ffn}.

\begin{equation}
\mathcal{F}_{\mathrm{out}} = \mathrm{FFN}(\mathcal{F}_{Fused}) .
\end{equation}

\subsection {Decoder Block}
\label{sec:decoder}
The Decoder Block manages to achieve a unitary processing of the fused features and provides the final representations after iteratively processing previous outputs concatenated with skip connections from the encoding path. 

In terms of the structure, it inherits the overall functionality of the Fusion Encoder Block, the only difference lying in the source of the Q, K, and V representations in its self-attention module and how the residual connections are formed in its structure.

Hence considering, $\mathcal{F}_{in} \in \mathbb{R}^{H \times W \times C}$, as the input feature and $\mathcal{F}_{out} \in \mathbb{R}^{H \times W \times C}$ as the resulting feature, following a plain ViT \cite{dosovitskiy2020image} structure, the underlying logic behind the Decoder Block can be formulated as:

\begin{gather}
\mathcal{F}'_{in} = \mathcal{F}_{in} + \text{LN}(\text{Attn}(\mathcal{F}_{in})), \\
\mathcal{F}_{out} = \mathcal{F}'_{in} + \text{LN}(\text{FFN}(\mathcal{F}'_{in})).
\end{gather}

The Q, K, and V representations are all derived from the same source, namely $\mathcal{F}_{\text{in}}$, in contrast to the separate main and auxiliary modality sources employed in the Fusion Self-Attention module. The input feature is first projected into a higher-dimensional space and then partitioned into three components corresponding to the attention inputs $Q$, $K$, and $V$.  

\begin{equation}
Q, K, V = \mathrm{Split}\big( \mathrm{Proj}(\mathcal{F}_{\text{in}}) \big), 
\quad 
Q, K, V \in \mathbb{R}^{H \times W \times C}.
\end{equation}

\subsection {Segmentation Head}
\label{sec:seg_head}
To map the high-dimensional fused representations extracted by the Fusion Vision Transformer into the distinct probability logits necessary for each class, a CNN is utilized.

Within the VesperMUSE architecture, the Segmentation Head utilizes a CNN to map the extracted feature representations into class logits, employing Batch-Normalisation (BN) to stabilize the activations. To act as a stochastic regularizer, a Dropout layer is strategically inserted between the two convolutional layers. By randomly zeroing hidden units during training, this layer prevents feature co-adaptation and forces the network to learn robust, distributed representations. Consequently, it mitigates overfitting to high-dimensional noise within the dense feature maps and ensures that the final pixel-wise classification generalizes effectively across diverse spatial contexts.

Let $cls$ denote the number of target semantic classes, $W_1$ and $W_2$ represent the weight matrices of the first and second convolutional layers respectively, and $\phi$ denote the Rectified Linear Unit (ReLU) activation function. Given an input feature map $\mathcal{F}_{\text{in}}$, the functionality of this Segmentation Head can thus be formalized as:

\begin{equation}
\mathrm{SegHead}(X; cls) 
= W_2 * \mathrm{Drop}\!\left(
\phi\!\left(
\mathrm{BN}\!\left(
W_1 * X
\right)
\right)
\right),
\quad
\mathrm{SegHead}(X; cls) \in \mathbb{R}^{H \times W \times cls}.
\end{equation}

\subsection {Fusion Vision Transformer Loss function}
In the datasets VesperMUSE tests semantic segmentation on, there is both a significant structural and semantic difference between the classes (especially on the CART \cite{lee2024cart} dataset) and also a very heavy class imbalance. To ensure multi-modal feature alignment and good class representation a weighted hybrid loss function is employed which also acts on multiple levels allowing deep supervision. 

Considering $\mathcal{L}_{\text{main}}(\cdot,\cdot)$ as the hybrid loss function, 
$\mathbf{P}^{(0)}$ as the final prediction, and 
$\mathbf{P}^{(s)}$ as the intermediate predictions originating from the decoding path and weighted by $\lambda_s$, 
with $\mathbf{Y}^{(0)}$ denoting the ground-truth labels and 
$\mathbf{Y}^{(s)}$ representing the ground-truth labels resized to match the spatial resolution of the corresponding deep supervision outputs, 
the total loss function $\mathcal{L}_{\text{total}}$ is formulated as follows:


\begin{equation}
\mathcal{L}_{\text{total}} 
=
\mathcal{L}_{\text{hybrid}}\!\left(\mathbf{P}^{(0)}, \mathbf{Y}\right)
+
\sum_{s \in \mathcal{S}}
\lambda_s \,
\mathcal{L}_{\text{hybrid}\!\left(\mathbf{P}^{(s)}, \mathbf{Y}^{(s)}\right)}.
\end{equation}


\noindent \textbf{Hybrid Loss. } $\mathcal{L}_{\text{hybrid}}(\cdot,\cdot)$ is a weighted combination 
of the multi-class Focal loss which operates at pixel level and is derived from the cross-entropy objective and the weighted Dice loss, which is region-based and directly optimizes the overlap between the predicted and ground-truth masks. Since the Dice loss is computed from class-wise statistics it is less sensitive to pixel-level imbalance and better captures structural and object-level coherence. 
The relative contribution of the constituents of the hybrid loss, $\mathcal{L}_{\text{hybrid}}$, is balanced by the coefficients $\alpha$ and $\beta$.

\begin{equation}
\mathcal{L}_{\text{hybrid}}
=
\alpha \, \mathcal{L}_{\text{Focal}}
+
\beta \, \mathcal{L}_{\text{Dice}}.
\end{equation}


\noindent \textbf{Multi-class Focal Loss. }
$\mathcal{L}_{\text{Focal}}(\cdot,\cdot)$ is a pixel-wise classification loss designed to mitigate severe class imbalance. 
The strong pixel-density of the \textit{background} class in both datasets can overwhelm the gradient signal and hinder learning of minority or hard classes. 
The Focal loss addresses this issue by introducing a modulating factor that reduces the importance of well-classified samples and emphasizes misclassified or uncertain predictions. 
Additionally, class-dependent weights are incorporated to further compensate for dataset imbalance.

Let $z_{i,c}$ denote the logit corresponding to class $c$ at pixel $i$, 
$C$ the total number of classes, and $p_{i,c}$ the softmax probability defined as:

\begin{equation}
p_{i,c}
=
\mathrm{softmax}(\mathbf{z}_i)_c
=
\frac{\exp(z_{i,c})}
{\sum_{k=0}^{C-1} \exp(z_{i,k})}.
\end{equation}

With $y_i \in \{0, \dots, C-1\}$ being the ground-truth class label for pixel $i$, 
$w_{y_i}$ the corresponding class weight, 
$\gamma$ the focusing parameter controlling the strength of down-weighting, 
and $N$ the total number of pixels, 
the multi-class Focal loss is formulated as:

\begin{equation}
\mathcal{L}_{\text{Focal}}
=
\frac{1}{N}
\sum_{i=1}^{N}
(1 - p_{i,y_i})^{\gamma}
\left(
- w_{y_i} \log(p_{i,y_i})
\right).
\end{equation}

\noindent \textbf{Weighted Dice Loss. }
$\mathcal{L}_{\text{Dice}}(\cdot,\cdot)$ is a region-based loss designed to directly optimize the overlap between the predicted segmentation and the ground-truth masks. 
Unlike pixel-wise classification losses, the Dice loss operates on global class-level statistics, making it inherently robust to class imbalance and particularly effective for small, sparse or weakly-represented objects. 
By incorporating class-dependent weights, the formulation further compensates for skewed class distributions while preserving structural consistency across object regions.

Let $p_{i,c}$ denote the predicted softmax probability for class $c$ at pixel $i$, 
$y_{i,c}$ the one-hot encoding of the ground-truth label, 
$N$ the total number of pixels, 
and $\epsilon$ a smoothing constant introduced for numerical stability. 
The Dice coefficient for class $c$ is defined as:

\begin{equation}
\mathrm{Dice}_c
=
\frac{
2 \sum_{i=1}^{N} p_{i,c} y_{i,c}
+
\epsilon
}
{
\sum_{i=1}^{N} p_{i,c}
+
\sum_{i=1}^{N} y_{i,c}
+
\epsilon
}.
\end{equation}

Let $w_c$ denote the class weight associated with class $c$, 
and $\delta$ a small constant preventing division by zero. 
Restricting normalization to classes present in the batch, 
the weighted Dice loss is expressed as:

\begin{equation}
\mathcal{L}_{\text{Dice}}
=
\frac{
\sum_{c=0}^{C-1}
w_c \left(1 - \mathrm{Dice}_c \right)
}
{
\sum_{c=0}^{C-1}
\mathbf{1}
\left(
\sum_{i=1}^{N} y_{i,c} > 0
\right)
+
\delta
}
\end{equation}.

\subsection {Fusion Vision Transformer Training}

The model is trained for 40K iterations for the PST900 dataset \cite{shivakumar2020pst900} and 80K iterations for the CART dataset \cite{lee2024cart} using the AdamW optimizer \cite{zhou2024towards} implemented in PyTorch \cite{paszke2019pytorch}, 
with a base learning rate of $3 \times 10^{-4}$, weight decay of $0.05$, 
and momentum coefficients $(\beta_1, \beta_2) = (0.9, 0.999)$. 
Gradient clipping with a maximum norm of $0.1$ is applied to stabilize optimization.

\noindent \textbf{Parameter-wise Learning Rates.}
Differential learning rates are employed for backbone and task-specific components. 
The ResNet backbones \cite{koonce2021resnet} are optimized with a learning rate scaled by $0.1$ relative to the base rate, while newly introduced fusion and prediction layers are trained with the full base learning rate. 
This strategy enables stable fine-tuning of pretrained representations while allowing faster convergence of randomly initialized layers.

\noindent \textbf{Learning Rate Scheduling.}
A cosine annealing schedule is employed with $T_{\max}=10K$ for the PST900 dataset \cite{shivakumar2020pst900} and $T_{\max}=20K$ for CART \cite{lee2024cart}. $T_{\max}$ controls the duration (in iterations) over which the learning rate 
gradually decreases from the initial value $\eta_0$ to the minimum value $\eta_{\min}$. The minimum learning rate  is
$\eta_{\min}=1 \times 10^{-7}$. A short warm-up phase is used.
Given that $\eta_0$ denotes the initial learning rate the learning rate at iteration $t$ is defined as:

\begin{equation}
\eta_t = \eta_{\min} 
+ \frac{1}{2}(\eta_0 - \eta_{\min})
\left(1 + \cos\left(\frac{\pi t}{T_{\max}}\right)\right).
\end{equation}

Cosine annealing is preferred because it provides a smooth, non-monotonic decay of the learning rate, 
allowing large updates in early stages for rapid convergence, followed by progressively smaller updates 
that improve stability and fine-grained parameter refinement. 
Compared to step decay, cosine annealing avoids abrupt learning rate drops, which can destabilize training.

For both semantic segmentation datasets the training is conducted on randomly-cropped patches. For PST900 \cite{shivakumar2020pst900} the dimension used is $512$x$512$, while on CART \cite{lee2024cart} it is $480$x$400$. 

\chapter{Experiments and Results}
\newpage

The evaluations for VesperMUSE across all tasks is conducted on two main datasets: PST900 \cite{shivakumar2020pst900} and CART \cite{lee2024cart}. The low-light adapted images and later enhanced are conducted on the same splits as used for the semantic segmentation evaluation.

\noindent \textbf{PST900.} The PST900 benchmark \cite{shivakumar2020pst900} consists of 894 synchronized RGB-Thermal image pairs acquired within demanding subterranean settings. The dataset features pixel-level semantic annotations for five distinct object categories: \textit{background, hand drill, backpack, fire extinguisher}, and \textit{survivor}. All modalities maintain a native resolution of $1280 \times 720$ pixels. For experimental consistency, the data is partitioned into training and evaluation subsets following a $2:1$ distribution ratio.

\noindent \textbf{CART.} The CART dataset \cite{lee2024cart} comprises 2,282 co-registered RGB-Thermal pairs captured across heterogeneous outdoor environments, including aquatic regions (rivers, lakes, coastlines), arid deserts, and dense forestation. Semantic labels are provided for a comprehensive 10-class taxonomy: \textit{bare ground, rocky terrain, developed structures, road, shrubs, trees, sky, water, vehicles}, and \textit{person}. The imagery is provided at a resolution of $960 \times 640$ pixels. The corpus is stochastically divided into training, validation, and testing sets using a $6:1:1$ ratio, respectively.

\textbf{Empirical contributions. } The results achieved by VesperMUSE across all levels and tasks can be summarised as:
\begin{itemize}
    \item \textbf{Robust Synthetic Low-Light Adaptation:} The synthetic adaptation of the PST900 \cite{shivakumar2020pst900} and CART \cite{lee2024cart} datasets achieves a superior SNR of approximately $10.7$ dB and a physical Gamma distribution of $1.00$. These metrics surpass established synthetic low-light adaptation pipelines such as the one used for creating LOL-v2 synthetic \cite{yang2021sparse} in terms of modeling low-light environments, ensuring that the low-light data maintains realistic luminance statistics and physics-informed consequences of light-deficiency.
    \item \textbf{Efficient and High-Performance LLIE:} The VesperMUSE LLIE component demonstrates significant performance improvements over existing methods, achieving a $29.5\%$ PSNR increase over LYT-Net \cite{lyt} and a $2.3\%$ increase over Retinexformer \cite{Retinexformer} on the synthetically-adapted CART dataset \cite{lee2024cart}. Notably, the model achieves performance comparable to the state-of-the-art ISALux \cite{balmez2026isalux} while utilizing only $0.454$M parameters, representing a $75.4\%$ reduction in model complexity.
    \item \textbf{State-of-the-Art Segmentation on CART:} On the CART dataset \cite{lee2024cart}, VesperMUSE establishes a new performance ceiling with an mIoU of $0.7672$, outperforming the top-performing CM-SSM \cite{guo2025cross} ($0.7460$) by $2.8\%$ and CMX \cite{zhang2023cmx} ($0.7400$) by $3.7\%$. Substantial class-wise gains are observed in challenging categories, including \textit{Vehicles} ($0.6314$) and \textit{Person} ($0.3426$), validating the effectiveness of thermal and structure-aware multi-modal fusion in outdoor environments.
    \item \textbf{Superior Generalization on PST900:} Evaluation on the PST900 dataset \cite{shivakumar2020pst900} 
yields a competitive mIoU of $0.8559$, representing a $1.4\%$ improvement over LASNet \cite{chen2022lasnet} and a $3.1\%$ increase over MDNet \cite{zhang2017mdnet}. The method sets new accuracy benchmarks for task-critical objects, outperforming CM-SSM \cite{guo2025cross} by $2.9\%$ for \textit{Hand-Drill} ($0.8282$) and by $2.4\%$ for \textit{Fire-Ext.} ($0.8776$), highlighting the framework's robustness across indoor dark scenarios.
\end{itemize}

\section{Synthetic Low-Light Image Adaptation}

The synthetic low-light adaptation pipeline described in Section \ref{sec:synthetic_ll_algo} must demonstrate robustness and reliability in accurately emulating real low-light conditions across all key factors that influence such environments, including illumination levels (both linear and non-linear/gamma-based), noise characteristics, inter-pixel relationships, and global statistical properties such as mean and standard deviation computed over the entire dataset.

To enable a rigorous quantitative comparison with existing datasets, encompassing both real and synthetic low-light imagery, a diverse set of evaluation metrics is employed. This comprehensive analysis ensures a thorough assessment of the effects induced by light deficiency and the fidelity of the proposed synthetic generation process. The proposed metrics are as follows:

\noindent\textbf{Signal-to-Noise Ratio (SNR).}
The goal is to quantify the level of corruption introduced by noise during the low-light generation process, while remaining invariant to global intensity scaling between the reference and degraded images. This is important because realistic low-light images are fundamentally characterized by photon-limited noise, and matching the real-world SNR distribution is critical for modeling the effect of low-light condition on the overall composition of the image. In the following formulation, $N$ denotes the reference normal-light image, $L$ the low-light image, $\mu_N$ and $\mu_L$ their mean intensities, $\alpha$ a scaling factor used to align global intensity levels, $\sigma_N$ the standard deviation of the noise residual, and $\Omega$ the image domain.

\begin{equation}
\mathrm{SNR} = 20 \log_{10} \left( \frac{\mu_{L}}{\sigma_{N}} \right), 
\quad
\sigma_{N} = \sqrt{\frac{1}{|\Omega|} \sum_{i \in \Omega} \left( L_i - \alpha N_i \right)^2}, 
\quad
\alpha = \frac{\mu_{L}}{\mu_{N}}.
\end{equation}

\noindent\textbf{Normalized Mutual Information (NMI).}
The objective is to measure how much structural information is preserved between the reference and the low-light image. Realistic degradation should alter illumination while maintaining scene structure, avoiding excessive distortion or artificial patterns. Given $N$ and $L$ denote the reference and degraded images, $p(x,y)$ the joint distribution of their pixel intensities, $p(x)$ and $p(y)$ the marginal distributions, and $H(\cdot)$ the entropy function, NMI can be formally formulated as:

\begin{equation}
\mathrm{NMI} = \frac{2 I(N,L)}{H(N) + H(L)}, 
\quad
I(N,L) = \sum_{x,y} p(x,y) \log \left( \frac{p(x,y)}{p(x)p(y)} \right).
\end{equation}

\noindent\textbf{Exposure Estimation.}
The aim is to capture the global illumination drop between the normal-light image and its low-light counterpart through the ratio of their average intensities. As the real low-light conditions are primarily governed by reduced exposure, and matching this factor ensures physically consistent brightness attenuation. This metric is represented as a ration between $\mu_N$ and $\mu_L$ representing the mean intensities of the reference and low-light images, respectively.

\begin{equation}
E = \frac{\mu_{L}}{\mu_{N}}.
\end{equation}

\noindent\textbf{Gamma Estimation.}
The purpose is to estimate the non-linear transformation applied to intensities during the low-light formation process by modeling the relationship between the reference and degraded image in logarithmic space. Given that $N_i$ and $L_i$ denote pixel intensities from the reference and low-light images, $E$ is the estimated exposure factor, $\epsilon$ is a small numerical constant for stability, and $\gamma$ represents the non-linearity coefficient, the non-linear (gamma) estimation can be expressed as. 

\begin{equation}
\gamma = \mathrm{median} \left( \frac{\log L_i}{\log (N_i \cdot E + \epsilon)} \right).
\end{equation}

\noindent\textbf{Mean Luminance.}
To characterize the overall brightness level of the synthesized dataset pixel intensities across the image domain are averaged. This is important because it provides a direct measure of how dark the generated images are, and offers a good insight in the environment conditions. In the following formulation, $Y_i$ denotes the luminance value of pixel $i$, and $\Omega$ represents the full image domain.

\begin{equation}
\mu_{Y} = \frac{1}{|\Omega|} \sum_{i \in \Omega} Y_i.
\end{equation}

\noindent\textbf{Luminance Standard Deviation.}
To quantify the spread of luminance values within an image, capturing intra-image contrast and variability is paramount. Realistic low-light scenes exhibit diverse illumination patterns rather than uniform darkness. To formally express this, $Y_i$ represents the luminance value at pixel $i$, $\mu_Y$ is the mean luminance, and $\Omega$ denotes the image domain.

\begin{equation}
\sigma_{Y} = \sqrt{ \frac{1}{|\Omega|} \sum_{i \in \Omega} (Y_i - \mu_{Y})^2 }.
\end{equation}

\noindent\textbf{Coefficient of Variation.}
The aim is to normalize luminance variability with respect to the mean brightness, providing a scale-invariant measure of illumination diversity across the dataset, and reflecting how well the synthetic data captures the variability present in real-world lighting conditions. In the following formulation, $\sigma_Y$ denotes the luminance standard deviation and $\mu_Y$ the mean luminance.

\begin{equation}
\mathrm{CV} = \frac{\sigma_{Y}}{\mu_{Y}} \times 100.
\end{equation}

\subsection{Quantitative Analysis}

\begin{table*}[t]
\centering
\small 
\setlength{\tabcolsep}{5pt} 
\begin{tabular}{@{} l c c c c c c c c @{}}
\toprule
\multirow{2}{*}{\textbf{Dataset}} & \multicolumn{2}{c}{\textbf{Quality Metrics}} & \multicolumn{3}{c}{\textbf{Image Statistics}} & \multicolumn{3}{c}{\textbf{Luminance (Y)}} \\
\cmidrule(lr){2-3} \cmidrule(lr){4-6} \cmidrule(lr){7-9} 
& \textbf{SNR (dB)} $\uparrow$ & \textbf{NMI} $\uparrow$ & \textbf{Exp.} $\downarrow$ & \textbf{Gamma} $\approx 1$ & \textbf{CV (\%)} $\uparrow$ & \textbf{Min / Max} & \textbf{Mean} & \textbf{Std} \\
\midrule
LOL v1 \cite{wei2018deep}           & 9.59  & 0.46 & 0.13 & 1.03 & 69.54 & 2.33 / 82.58  & 15.49 & 10.77 \\
LOL v2 real \cite{yang2021sparse}      & 8.51  & 0.34 & 0.15 & 1.02 & 69.34 & 0.34 / 82.58  & 16.08 & 11.16 \\
LOL v2 synthetic \cite{yang2021sparse}     & 4.73  & 0.60 & 0.46 & 1.38 & 56.97 & 8.26 / 186.97 & 61.90 & 35.27 \\
LOL Blur \cite{LOL-Blur}         & 8.70  & 0.47 & 0.28 & 1.07 & 48.98 & 4.48 / 70.34  & 27.23 & 13.34 \\
\midrule
PST900 LL & \textbf{10.71} & 0.31 & 0.29 & \textbf{1.00} & 51.96 & 3.38 / 29.59  & 12.86 & 6.68  \\
CART LL & \textbf{10.73} & 0.29 & 0.16 & \textbf{1.00} & 54.40 & 3.38 / 36.73  & 12.71 & 6.91  \\
\bottomrule
\end{tabular}
\vspace{2mm} 
\caption{Quantitative comparison between real and  low-light datasets across multiple statistical and physical metrics. The PST900 \cite{shivakumar2020pst900}/CART \cite{lee2024cart} synthetic  table entries are paired images comprised from original RGB images from the two datasets and synthetically adapted low-light images.}
\label{tab:low_light_comparison}
\end{table*}

The purpose of VesperMUSE's proposed synthetic adaptation algorithm was to represent the synthetic low-light images as close to the real-world low-light imaging present in LOL-v1 \cite{wei2018deep}, LOL-v2-real \cite{yang2021sparse} and LOL Blur \cite{LOL-Blur} as well as improve on other methods which proposed synthetic adaptation methods such as LOL-v2-synthetic \cite{yang2021sparse}.

\textbf{Illumination and Intensity. } To accurately assess the perceptual quality and light profile from the synthetically adapted images, it is required to interpret the absolute intensity, the non-linearity in the captured illumination as well as uniformity of the light distribution. The Mean Luminance assesses the physical brightness in the scene which is observed to be low in the real world datasets LOL-v1 \cite{wei2018deep} ($15.49$) and LOL-v2-real \cite{yang2021sparse} ($16.08$), but almost $4$ times higher in LOL-v2-synthetic \cite{yang2021sparse} ($61.90$). These values are strongly linked with the exposure which acts as a linear scalar multiplier for the incoming irradiance. The datasets obtained with the proposed method manage to share more similarity with the real-captured datasets in both mean luminance and exposure.

Moreover, a critical flaw in earlier synthetic approaches, is the introduction of artificial non-linearity values, with Gamma being on average approximately $26\%$ higher than the values observed in real-world scenarios. In contrast, the proposed pipeline demonstrates a perfectly linear mapping, much closer to other low-light imaging datasets conditions.

\begin{figure}[t]
    \centering
    \includegraphics[width=0.7\linewidth]{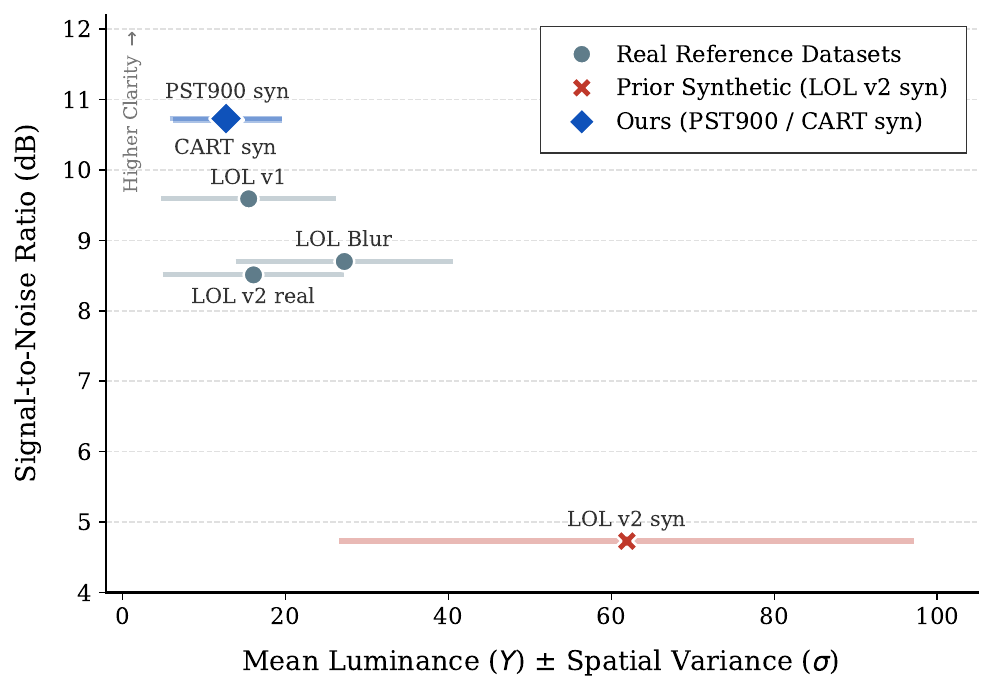}
    \vspace{-2mm} 
    \caption{Illumination vs. signal integrity. Real low-light data retain high SNR, while prior synthetic methods introduce excessive noise and over-brightening. VesperMUSE preserves an accurate signal quality similar to that observable in true low-light conditions.}
    \label{fig:luminance_snr}
\end{figure}

\textbf{Noise. } A well-known problem of light-deficient scenes is the presence of noise due to the reduced frequency of photons captured by the sensor. Therefore, an accurate noise modeling must always be linked with the level of illumination. The level of alignment with the real world datasets as well as the coherence of this correlation between the level of light and noise can be visualised in Figure \ref{fig:luminance_snr}.

\textbf{Information preservation. } Both local and global structural information is also lost when comparing a low-light image and its well-lit counterpart. The absence of light naturally introduces colour distortions and more weak edges which ultimately lead to overall information loss which is measured using the NMI. The two proposed datasets present the most important challenge in terms of the amount of information that needs to be recovered through the LLIE process, mainly when compared with the LOL-v2-synthetic \cite{yang2021sparse} where the shared information is much higher between the low-light and normal light images ($0.60$ compared with $0.31$ and $0.29$ respectively). 

\subsection{Qualitative Analysis}

\begin{figure}[t]
    \centering
    \includegraphics[width=\linewidth]{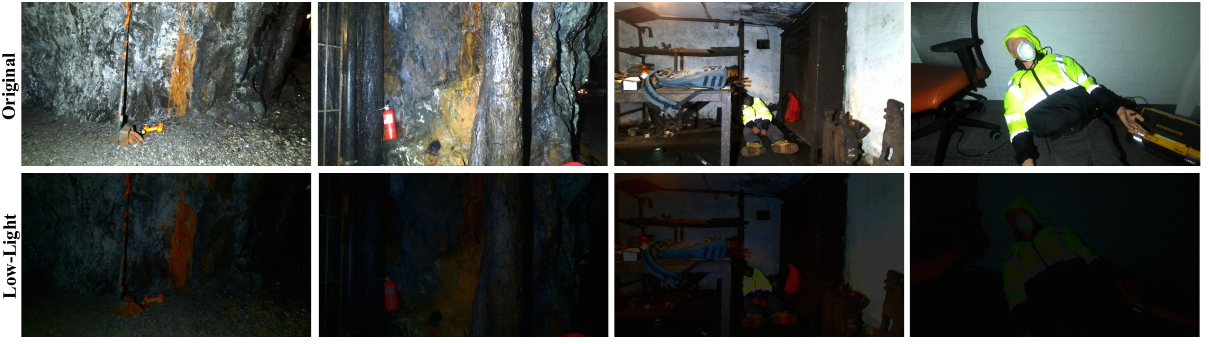}
    \vspace{-2mm} 
    \caption{PST900 \cite{shivakumar2020pst900} synthetic low-light adapted RGB and the original image.}
    \label{fig:pst900_syn_ll_qualitative}
\end{figure}

\begin{figure}[t]
    \centering
    \includegraphics[width=\linewidth]{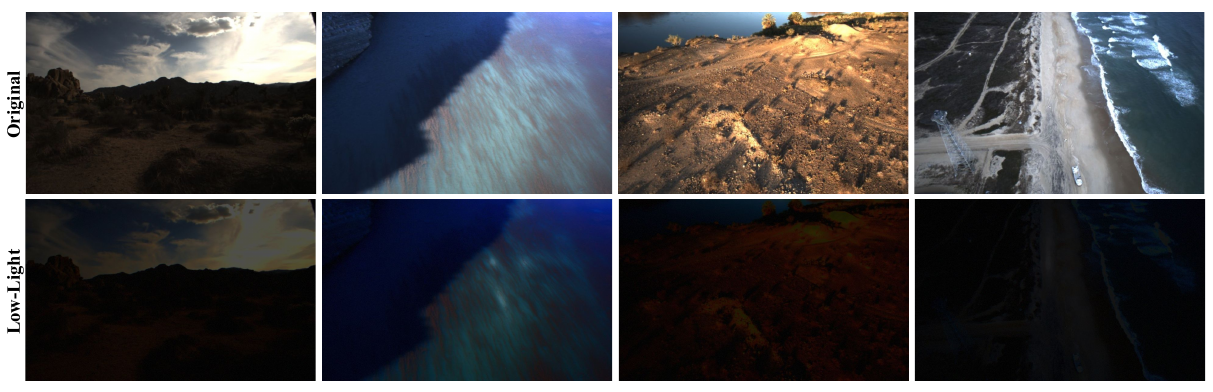}
    \vspace{-2mm} 
    \caption{CART \cite{lee2024cart} synthetic low-light adapted RGB and the original image.}
    \label{fig:cart_syn_ll_qualitative}
\end{figure}

In terms of perceptual image quality, VesperMUSE generates a low-light counterpart that degrades the input in accordance with characteristic low-light conditions. Notably, colour distortions and structured noise patterns are clearly observable in the synthetically generated samples (Figures \ref{fig:pst900_syn_ll_qualitative}, \ref{fig:cart_syn_ll_qualitative}). Furthermore, the illumination has spatially and statistically a high variability, similar to the real-world environmental conditions.

\section{Low-Light Image Enhancement}
\textbf{Evaluation Metrics.} To quantitatively evaluate the fidelity of the reconstructed images, PSNR is the main utilized measure. This metric is defined by the ratio between the maximum possible power of a signal and the power of corrupting noise that affects the fidelity of its representation. In this formulation, $M$ and $N$ represent the image dimensions, $I(i,j)$ and $K(i,j)$ denote the pixel intensities of the ground truth and reconstructed images at coordinates $(i,j)$, and $MAX_I$ signifies the maximum possible pixel value. The PSNR is expressed as:
\begin{equation}
\text{PSNR} = 10 \cdot \log_{10} \left( \frac{MAX_I^2}{\frac{1}{MN} \sum_{i=0}^{M-1} \sum_{j=0}^{N-1} [I(i,j) - K(i,j)]^2} \right).
\end{equation}

Complementary to the pixel-level error captured by PSNR, Structural Similarity Index Measure (SSIM) \cite{ssim} is also employed to better account for perceived visual quality and structural integrity. While PSNR relies on absolute differences, SSIM evaluates the similarity between the ground truth and the output by comparing luminance, contrast, and structural information across two images $x$ and $y$. In the following expression, $\mu_x$ and $\mu_y$ represent the local means, $\sigma_x^2$ and $\sigma_y^2$ denote the local variances, $\sigma_{xy}$ signifies the cross-covariance, and $c_1$ and $c_2$ are constants incorporated to ensure numerical stability, the formal representation of SSIM can be expressed as:
\begin{equation}
\text{SSIM}(x, y) = \frac{(2\mu_x\mu_y + c_1)(2\sigma_{xy} + c_2)}{(\mu_x^2 + \mu_y^2 + c_1)(\sigma_x^2 + \sigma_y^2 + c_2)}.
\end{equation}

\subsection{Quantitative Analysis}

\begin{table*}[ht]
\centering
\resizebox{0.7\textwidth}{!}{%
\begin{tabular}{l ccc ccc c}
\toprule
\multirow{2}{*}{\textbf{Method}} & \multicolumn{2}{c}{\textbf{PST900 LL}} & \multicolumn{2}{c}{\textbf{CART LL}} & \multirow{2}{*}{\textbf{Params (M) $\downarrow$}} \\
\cmidrule(lr){2-3} \cmidrule(lr){4-5}
 & PSNR $\uparrow$ & SSIM $\uparrow$ & PSNR $\uparrow$ & SSIM $\uparrow$ & \\
\midrule
LYT-Net \cite{lyt} & 22.81 & 0.860 & 21.42 & 0.832 & \textbf{0.045} \\
Retinexformer \cite{Retinexformer} & 26.60 & 0.850 & 27.11 & 0.847 & 1.61 \\
ISALux \cite{balmez2026isalux} & \textbf{27.58} & \textbf{0.874} & \textbf{27.81} & \textbf{0.873} & 1.85 \\
\midrule
\textbf{VesperMUSE LLIE} & 27.47 & 0.870 & 27.74 & 0.858 & 0.454 \\
\bottomrule
\end{tabular}%
}
\vspace{0.5em}
\caption{Quantitative comparison of LLIE methods on the synthetically low-light adapted PST900 LL and CART LL datasets. Best results are highlighted in \textbf{bold}.}
\label{tab:llie_comparison}
\end{table*}

The quantitative results in Table~\ref{tab:llie_comparison} highlight several important points about the behaviour of thermal-assisted enhancement. An observation that needs to be made is that VesperMUSE LLIE is natively thermal-aware, while the other methods only aim to predict the well-lit images from the low-light maps only. ISALux \cite {balmez2026isalux} achieves the highest overall performance across both datasets, reaching $27.58$\,dB/$0.874$ (PSNR/SSIM) on PST900 LL and $27.81$\,dB/$0.873$ on CART LL, indicating strong reconstruction fidelity under purely RGB-based enhancement. However, VesperMUSE LLIE demonstrates highly competitive performance, attaining $27.47$\,dB/$0.870$ on PST900 LL and $27.74$\,dB/$0.858$ on CART LL, while maintaining a substantially lower parameter count of $0.454$M compared to $1.85$M. This corresponds to approximately a $4\times$ reduction in model size with only marginal degradation in PSNR ($\leq 0.11$\,dB) and a limited drop in SSIM, particularly on PST900 LL.

Compared to Retinexformer \cite{Retinexformer}, VesperMUSE LLIE consistently improves PSNR on both datasets (+$0.87$\,dB on PST900 LL and +$0.63$\,dB on CART LL), while using approximately $3.5\times$ fewer parameters, highlighting a significantly better performance-to-complexity trade-off. Although Retinexformer \cite{Retinexformer} exhibits slightly higher SSIM on CART LL, but the gap remains small, suggesting that structural preservation is largely maintained by the proposed method.

Notably, the strong performance of VesperMUSE LLIE can be attributed to its thermal-aware design, which introduces complementary modality information that is particularly beneficial under challenging low-light conditions where RGB signals alone are not always stable and informative. This is especially evident on PST900 LL, where the gap to the top-performing method is minimal despite architectural differences and reduced model capacity. In contrast, LYT-Net \cite{lyt}, while extremely lightweight ($0.045$M parameters), suffers from a significant drop in both PSNR and SSIM, indicating that aggressive parameter reduction without multi-modal guidance limits restoration quality.

\subsection{Qualitative Analysis}

The qualitative results further demonstrate the strong enhancement capability of VesperMUSE LLIE, particularly in approximating the RGB reference from severely degraded inputs. The proposed method produces outputs with improved local contrast and more accurate colour saturation compared to Retinexformer \cite{Retinexformer} and LYT-Net \cite{lyt}, which tend to generate flatter or slightly desaturated images. Notably, chromatic fidelity is well preserved, as observed in the PST900 LL samples where objects such as the yellow helmet and red fire extinguisher are reconstructed with more natural and vivid colours.

The benefits of the thermal-aware design are also evident in the preservation of structural details. In regions where the low-light RGB input lacks visible information, VesperMUSE LLIE effectively leverages thermal cues to enhance edge sharpness and suppress noise, leading to clearer object boundaries and more stable textures. While ISALux \cite{balmez2026isalux} achieves comparable visual quality, the proposed method delivers competitive results with a significantly lower computational footprint, avoiding common artifacts such as halos or over-smoothing.

\begin{figure}[H]
    \centering
    \includegraphics[width=\linewidth]{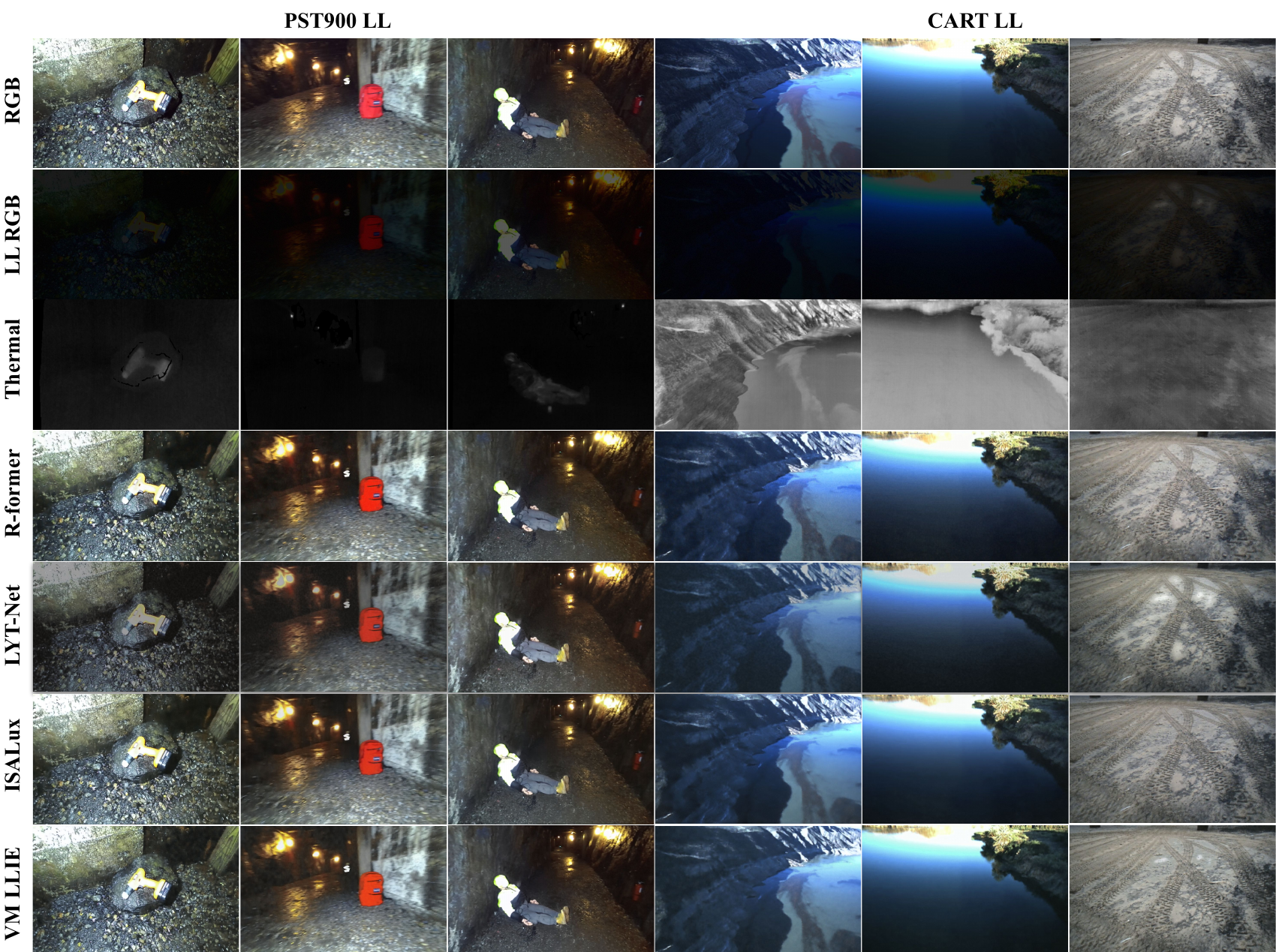}
    \vspace{-2mm} 
    \caption{Qualitative comparison of VesperMUSE (VM) LLIE with SOTA methods enhancing the synthetically low-light adapted PST900 LL and CART LL datasets.}
    \label{fig:llie_qualitative}
\end{figure}

\section{Semantic Segmentation}

\noindent \textbf{Evaluation Metrics.} To quantitatively evaluate the efficacy of the proposed semantic segmentation framework, we employ the mIoU  as the primary performance metric. Given that  $N$ denotes the total number of semantic classes, and $TP$, $FP$, and $FN$ represent the true positives, false positives, and false negatives for class $i$, respectively, the mIoU can be formalised as:
\begin{equation}
    \text{mIoU} = \frac{1}{N} \sum_{i=1}^{N} \frac{TP_i}{TP_i + FP_i + FN_i}.
\end{equation}

\subsection{Quantitative Analysis}

\textbf{PST900 Evaluation.} While VesperMUSE achieves a highly competitive mIoU of $0.8559$ (trailing the leading CM-SSM \cite{guo2025cross} by a marginal $0.0031$), a closer examination of class-wise distributions reveals the distinct advantage of its tri-modal input architecture. Despite employing a comparatively standard ResNet50/18 \cite{koonce2021resnet} backbone ensemble, the model establishes new state-of-the-art thresholds for geometrically rigid, contextually isolated objects. It notably outperforms robust contemporary baselines, including SA-Gate \cite{chen2020bi} and CM-SSM \cite{guo2025cross}, on the \textit{Hand-Drill} ($0.8282$) and \textit{Fire-Extinguisher} ($0.8776$) classes by significant margins. This suggests that the inclusion of surface normal maps act as an informative structural guide, allowing the network to delineate sharp object boundaries even against methods utilizing substantially heavier feature extractors (e.g., LASNet's ResNet-152 \cite{koonce2021resnet}). Furthermore, the VesperMUSE LLIE configuration demonstrates remarkable robustness, achieving a strong $0.7818$ mIoU that remains highly competitive with established baselines such as EGFNet \cite{dong2023egfnet}. Therefore, it is shown that the pre-enhancement process aids semantic understanding through noise removal and colour shift correction (two essential components of low-light information recovery) while simultaneously producing visually appealing, illuminated feeds.

\begin{table*}[ht]
\centering
\resizebox{\textwidth}{!}{%
\begin{tabular}{ll ccccc c}
\toprule
\multirow{2}{*}{\textbf{Method}} & \multirow{2}{*}{\textbf{Backbone}} & \multicolumn{5}{c}{\textbf{IoU per Class (\%) $\uparrow$}} & \multirow{2}{*}{\textbf{mIoU (\%) $\uparrow$}} \\
\cmidrule(lr){3-7}
 & & Background & Hand-Drill & Backpack & Fire-Ext. & Survivor & \\
\midrule
EGFNet \cite{dong2023egfnet} & - & 99.26 & 64.67 & 83.05 & 71.29 & 74.30 & 78.51 \\
SA-Gate \cite{chen2020bi} & - & 99.25 & 81.01 & 79.77 & 72.97 & 62.22 & 79.05 \\
MMSMCNet$_{23}$ \cite{zhou2023mmsmcnet} & Mit-B3 & - & 62.40 & 89.20 & 73.30 & 74.70 & 79.80 \\
CLNet-T$_{24}$ \cite{guo2024contrastive} & Mit-B4 & - & 68.50 & 87.90 & 72.70 & 75.70 & 80.80 \\
DBCNet$_{23}$ \cite{zhou2023dbcnet} & ResNet101 & - & 77.20 & 82.70 & 73.00 & 76.70 & 81.80 \\
C$^4$Net-Mit$_{25}$ \cite{zhao2024c4net} & Mit-B2 & - & 69.40 & 88.40 & 79.60 & 79.60 & 82.60 \\
SGFNet$_{23}$ \cite{wang2023sgfnet} & ResNet152 & - & 76.70 & 85.40 & 75.60 & 76.70 & 82.80 \\
MDNet$_{25}$ \cite{zhang2017mdnet} & Mit-B2 & - & 69.50 & \textbf{89.40} & 76.80 & 80.20 & 83.00 \\
LASNet \cite{chen2022lasnet} & ResNet-152 & \textbf{99.46} & 77.75 & 86.48 & 82.80 & \textbf{83.43} & 84.40 \\
CM-SSM \cite{guo2025cross} & EfficientVit-B1 & - & 80.50 & 85.40 & 85.70 & 78.10 & \textbf{85.90} \\
\midrule
\textbf{VesperMUSE LLIE} & \textbf{ResNet50/ResNet18} & 99.00 & 75.73 & 56.89 & 82.08 & 77.22 & 78.18 \\
\textbf{VesperMUSE} & \textbf{ResNet50/ResNet18} & \textbf{99.46} & \textbf{82.82} & 83.36 & \textbf{87.76} & 74.56 & 85.59 \\
\bottomrule
\end{tabular}%
}
\vspace{0.5em} 
\caption{Performance comparison with SOTA methods on the PST900 \cite{shivakumar2020pst900} dataset. Best results are highlighted in \textbf{bold}.}
\label{tab:sota_pst900_backbones}
\end{table*}

\textbf{CART Evaluation.} VesperMUSE establishes a new state-of-the-art on the operationally complex CART dataset \cite{lee2024cart}, achieving an mIoU of $0.7672$ and decisively outperforming previous leading methods such as CM-SSM \cite{guo2025cross} ($0.7460$) and CMX \cite{zhang2023cmx} ($0.7400$). In contrast to PST900 \cite{shivakumar2020pst900}, CART \cite{lee2024cart} necessitates pronounced resilience to unstructured outdoor environments and highly variable terrain boundaries. The architecture's efficacy on ground-level classifications (e.g., \textit{Bare Grd.} at $0.8706$ compared to CM-SSM's \cite{guo2025cross} $0.8260$, and \textit{Road} at $0.6538$ over CRM's \cite{shin2024complementary} $0.6160$) underscores the robust spatial synergy between thermal features and surface normals. This cross-modal integration effectively compensates for the severe degradation of visual texture typical of low-light outdoor scenes. Critically, VesperMUSE exhibits superior semantic preservation of dynamic obstacles, surpassing recent baselines like MCNet-T \cite{guo2025transferring} on the highly challenging \textit{Person} class ($0.3426$ vs. $0.3150$) and providing a new best mIoU on \textit{Vehicles} class with $0.6314$. Conversely, while the VesperMUSE LLIE variant encounters challenges with fine-grained dynamic obstacles (most notably its inability to delineate the \textit{Person} class ($0.0000$)), this explicit LLIE configuration offers two distinct operational advantages. As opposed to the fine-features constructed classes, the enhancement process successfully reconstructs macro-level environmental features, maintaining robust segmentation accuracy on foundational terrain classes such as \textit{Water} ($0.8996$) and \textit{Rocky Terr.} ($0.7891$).

\begin{table*}[ht]
\centering
\resizebox{\textwidth}{!}{%
\begin{tabular}{ll cccccccccc c}
\toprule
\multirow{2}{*}{\textbf{Model}} & \multirow{2}{*}{\textbf{Backbone}} & \multicolumn{10}{c}{\textbf{IoU per Class (\%) $\uparrow$}} & \multirow{2}{*}{\textbf{mIoU (\%) $\uparrow$}} \\
\cmidrule(lr){3-12}
 & & Bare Grd. & Rocky Terr. & Dev. Struct. & Road & Shrubs & Trees & Sky & Water & Vehicles & Person & \\
\midrule
GMNet$_{21}$ \cite{zhou2021gmnet} & ResNet152 & 81.50 & 88.80 & 83.80 & 60.30 & 70.80 & 79.50 & 92.00 & 97.40 & 54.80 & 17.80 & 72.70 \\
EGFNet$_{22}$ \cite{dong2023egfnet} & ResNet101 & 80.30 & 85.20 & 56.30 & 60.50 & 69.90 & 78.40 & 92.80 & 97.40 & 57.20 & 17.10 & 69.50 \\
EAEFNet$_{23}$ \cite{liang2023explicit} & ResNet152 & 80.50 & 86.30 & 81.30 & 54.50 & 70.20 & \textbf{80.20} & \textbf{93.70} & 97.60 & 48.10 & 0.00 & 69.20 \\
CMNext$_{23}$ \cite{zhang2023delivering} & Mit-B2 & 80.10 & 87.80 & 84.10 & 58.60 & 69.90 & 78.50 & 92.50 & 97.30 & 54.30 & 17.80 & 72.10 \\
CMX$_{23}$ \cite{zhang2023cmx} & Mit-B2 & 82.40 & 90.30 & 85.00 & 60.40 & 71.00 & 79.90 & 93.60 & 97.60 & 57.90 & 22.20 & 74.00 \\
CRM$_{24}$ \cite{shin2024complementary} & Swin-B & 81.60 & 90.00 & 85.10 & 61.60 & 69.50 & 77.90 & 92.30 & 97.40 & 37.70 & 20.60 & 71.40 \\
CLNet-T$_{24}$ \cite{guo2024contrastive} & Mit-B4 & 81.80 & 89.30 & 84.80 & 61.50 & 71.30 & 79.50 & 93.40 & 97.50 & 52.30 & 25.10 & 73.70 \\
CAINet$_{24}$ \cite{li2024coupled} & MobileNetV2 & 80.20 & 88.40 & 78.40 & 57.10 & 68.80 & 77.00 & 92.10 & 96.90 & 43.90 & 17.20 & 70.00 \\
MCNet-T$_{25}$ \cite{guo2025transferring}  & ConvNeXtV2-B & 80.50 & 86.90 & 83.70 & 58.00 & 69.70 & 78.00 & 92.90 & 97.70 & 58.10 & 31.50 & 73.70 \\
CM-SSM \cite{guo2025cross} & EfficientVit-B1 & 82.60 & 89.70 & 84.90 & 61.00 & 71.30 & 79.10 & 92.90 & 97.60 & 62.00 & 24.50 & 74.60 \\
\midrule
\textbf{VesperMUSE LLIE} & \textbf{ResNet50/ResNet18} & 75.44 & 78.91 & 65.90 & 48.25 & 57.87 & 65.43 & 76.69 & 89.96 & 37.20 & 0.00 & 59.57 \\
\textbf{VesperMUSE} & \textbf{ResNet50/ResNet18} & \textbf{87.06} & \textbf{90.77} & \textbf{86.72} & \textbf{65.38} & \textbf{72.57} & 77.98 & 91.50 & \textbf{97.78} & \textbf{63.14} & \textbf{34.26} & \textbf{76.72} \\
\bottomrule
\end{tabular}%
}
\vspace{0.5em}
\caption{Performance comparison with SOTA methods on the CART dataset \cite{lee2024cart}. Best results are highlighted in \textbf{bold}.}
\label{tab:sota_cart}
\end{table*}

\subsection{Qualitative Analysis}

\textbf{PST900 Evaluation. } VesperMUSE's strong quantitative results in underground scenarios are reinforced by visual comparisons against HAFNet \cite{peng2023hafnet}, CRM-RGBTSeg \cite{shin2024complementary}, CAINet \cite{li2024coupled}, and EAEFNet \cite{liang2023explicit}. First, the standard VesperMUSE configuration consistently maintains crisp, accurate predictions for primary targets, such as humans and equipment. By effectively fusing thermal and RGB data, VesperMUSE isolates heat-emitting objects even in near-total darkness, avoiding the boundary fragmentation seen in baselines like CAINet \cite{li2024coupled}. 

\begin{figure}[H]
    \centering
    \begin{minipage}[b]{0.48\linewidth} 
        \centering
        \includegraphics[width=\linewidth]{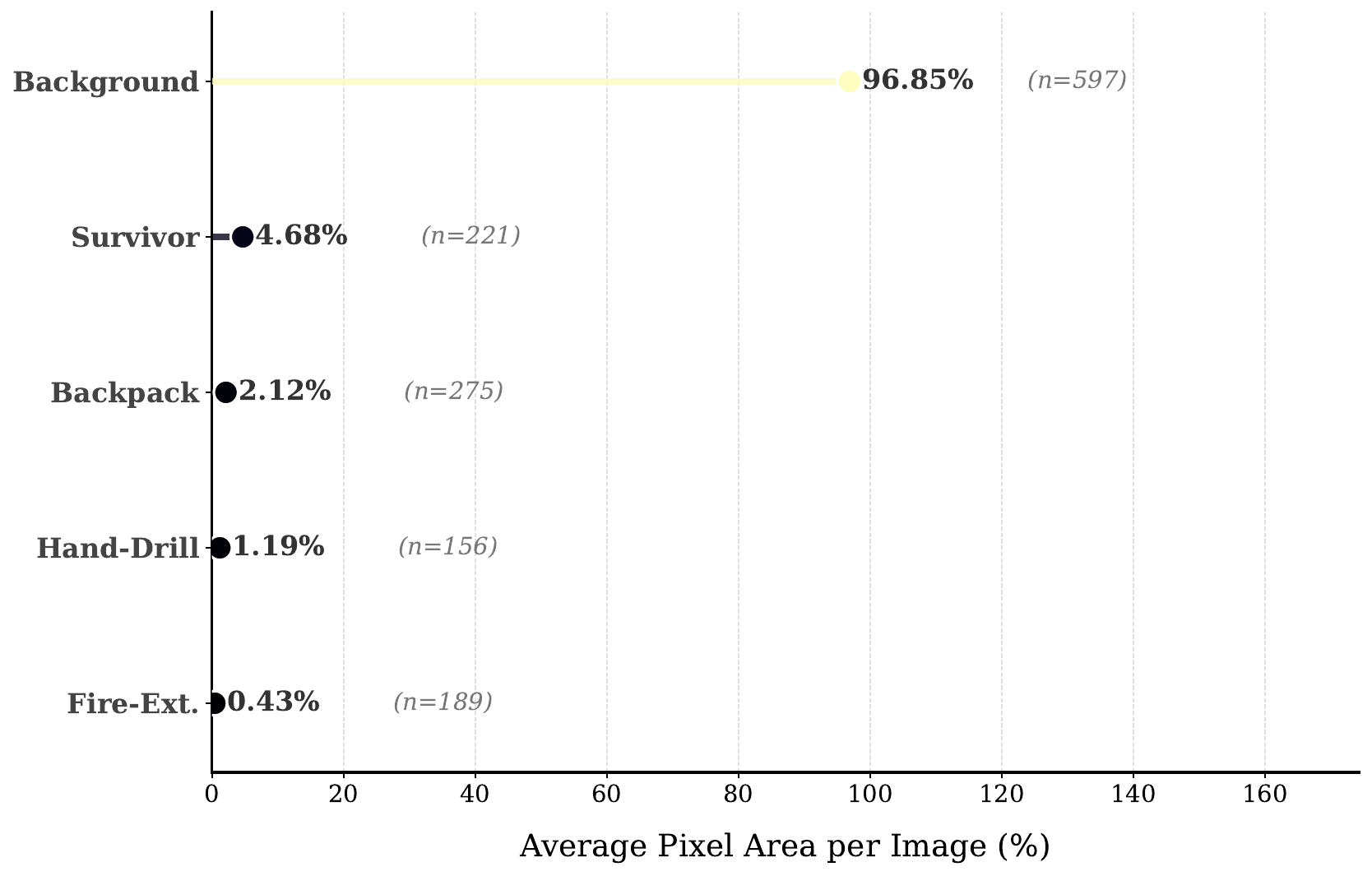}
        \vspace{-2mm}
    \end{minipage}\hfill
    \begin{minipage}[b]{0.48\linewidth} 
        \centering
        \includegraphics[width=\linewidth]{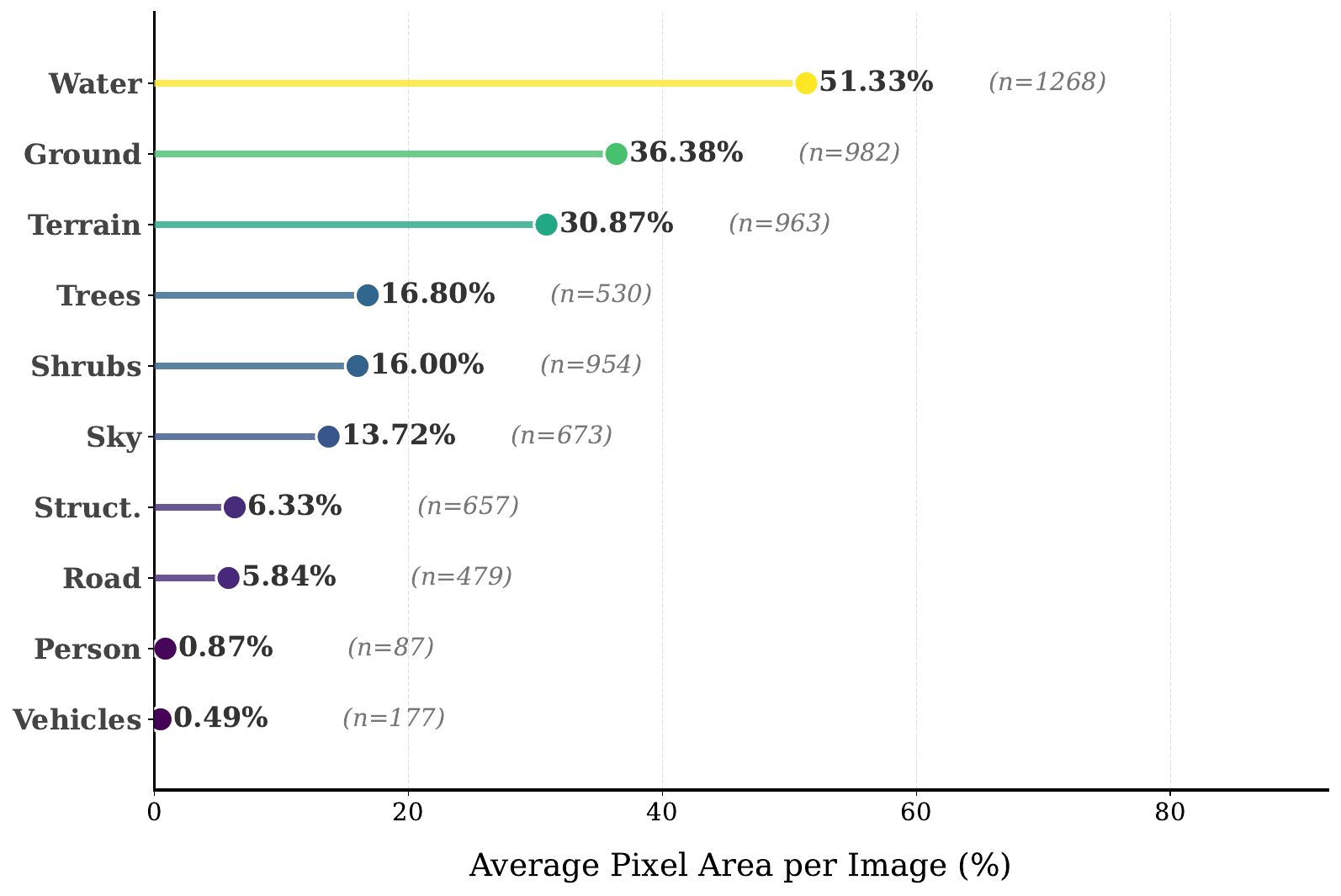}
        \vspace{-2mm}
    \end{minipage}
    
    \caption{Illustration of the per-class surface distribution on the PST900 \cite{shivakumar2020pst900} (left) and CART \cite{lee2024cart} (right) datasets. The percentage for each class is computed as the mean percentage of pixels covered across the samples where it appears (sample count is encoded as $n$). }
    \label{fig:surface_distributions_combined}
\end{figure}

\begin{figure}[H]
    \centering
    \includegraphics[width=\linewidth]{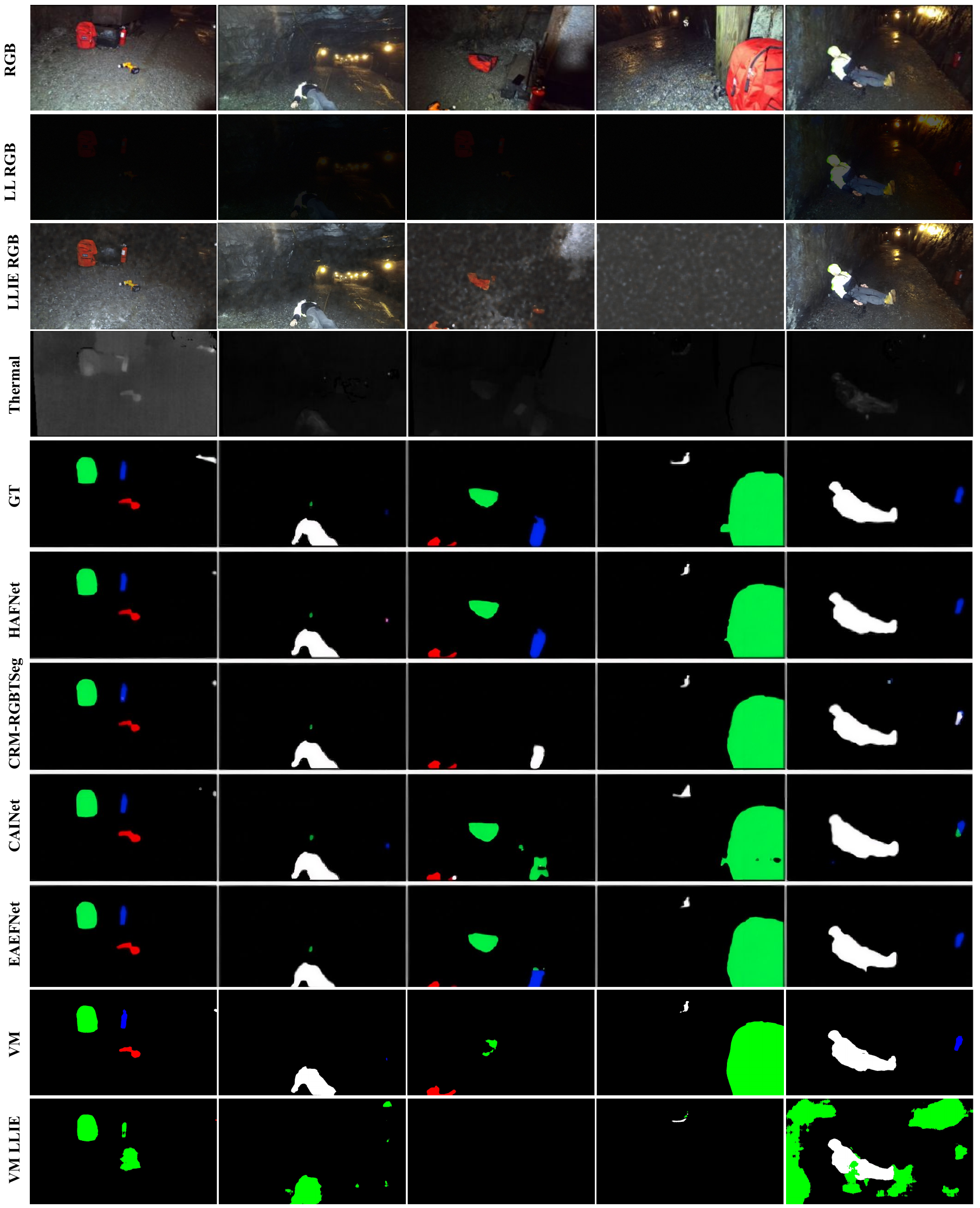}
    \vspace{-2mm} 
    \caption{Qualitative comparison of VesperMUSE (VM) and the LLIE configuration (VM LLIE) with SOTA models on the PST900 dataset \cite{shivakumar2020pst900}.}
    \label{fig:pst_seg_qualitative}
\end{figure}

Furthermore, the VesperMUSE LLIE configuration demonstrates a distinct advantage in pitch-black environments. While the noise is still a very serious problem and some classes with objects covering a smaller surface get misclassified because of the low-level feature shift caused by low-light environments, the LLIE configuration manages to recover essential hidden structural details even when surrounded by heavy noise (the \textit{Survivor} instance on the last column of Figure \ref{fig:pst_seg_qualitative}).

\textbf{CART Evaluation. } The strong quantitative performance on CART \cite{lee2024cart} is supported by VesperMUSE's visually comparison against methods like SFAF-MA \cite{he2023sfaf}, CMNext \cite{zhang2023delivering}, SGFNet \cite{wang2023sgfnet}, CM-SSM \cite{guo2025cross}. A first observation would be that the proposed method manages to continuously maintain a prediction for larger objects and not predict other classes inside of them (this behaviour can be noticed for SFAF-MA \cite{he2023sfaf} and CMNext \cite{zhang2023delivering}).

\begin{figure}[H]
    \centering
    \includegraphics[width=\linewidth]{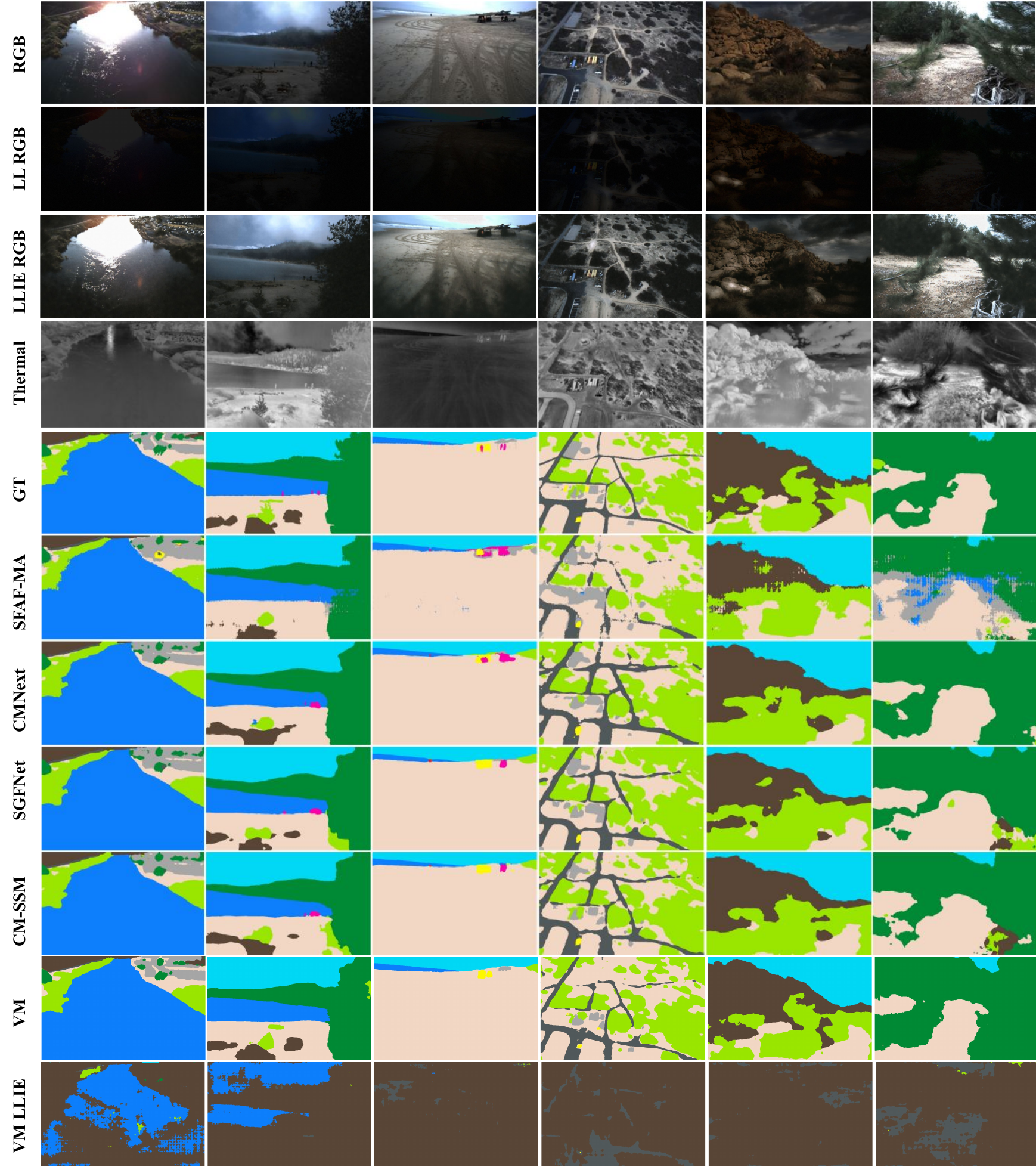}
    \vspace{-2mm} 
    \caption{Qualitative comparison of VesperMUSE (VM) and the LLIE configuration (VM LLIE) with SOTA models on the CART dataset.}
    \label{fig:cart_seg_qualitative}
\end{figure}

Moreover, the last column of Figure \ref{fig:cart_seg_qualitative} shows another advantage of VesperMUSE on three out of the 4 compared models. In environments where multiple classes intertwine, all methods besides VesperMUSE and CMNext \cite{zhang2023delivering} seem to get confused on which is the dominant class in the problematic regions, as the \textit{Trees} class is often erroneously conflated with \textit{Rocky Terrain} or \textit{Bare Ground}. By taking a look at the qualitative results for the VesperMUSE LLIE configuration, it can be easily concluded that the performance comes from predominant classes with very large objects like: \textit{Bare Ground}, \textit{Rocky Terrain}, \textit{Water}, and that correctly segmenting smaller classes for the majority of the samples still remains a challenge.

\section{Computational Performance}

\begin{table}[ht]
\centering
\resizebox{0.7\linewidth}{!}{%
\begin{tabular}{ll ccc}
\toprule
\textbf{Model} & \textbf{Backbone} & \textbf{FLOPs (G) $\downarrow$} & \textbf{Params (M) $\downarrow$} & \textbf{mIoU (\%) $\uparrow$} \\
\midrule
SGFNet$_{23}$ \cite{wang2023sgfnet} & ResNet152 & 143.73 & 125.25 & 68.5 \\
CMNeXt$_{23}$ \cite{zhang2023delivering} & Mit-B2 & 68.70 & 58.68 & 72.1 \\
CMX$_{23}$ \cite{zhang2023cmx} & Mit-B2 & 67.20 & 66.57 & 74.0 \\
CLNet-T$_{24}$ \cite{guo2024contrastive} & Mit-B4 & 217.85 & 130.84 & 73.7 \\
CAINet$_{24}$ \cite{li2024coupled} & MobileNetV2 & 123.62 & \textbf{12.16} & 70.0 \\
MCNet-T$_{25}$ \cite{guo2025transferring} & ConvNeXt-B & 278.85 & 199.79 & 73.7 \\
CM-SSM \cite{guo2025cross} & EfficientVit-B1 & \textbf{10.34} & 12.59 & 74.6 \\
\midrule
\textbf{VesperMUSE} & \textbf{ResNet50/ ResNet18} & 321.28 & 180.45 & \textbf{76.7} \\
\bottomrule
\end{tabular}%
}
\vspace{0.5em}
\caption{Complexity comparison with an image resolution of $480 \times 640$ pixels. mIoU results are reported on the CART dataset \cite{lee2024cart}.}
\label{tab:computational_complexity}
\end{table}

Although the multi-modal architecture of VesperMUSE inherently incurs a higher computational cost in terms of floating-point operations (FLOPs), it maintains a competitive parameter footprint relative to its performance capabilities. Specifically, VesperMUSE establishes a new SOTA on the CART dataset \cite{lee2024cart} with an mIoU of $0.767$, outperforming the previous leading method (CM-SSM \cite{guo2025cross}) by a significant margin of $2.1$\%. Furthermore, when compared to other larger architectures such as MCNet-T \cite{guo2025transferring}, VesperMUSE delivers a $3.0$\% absolute improvement in mIoU while utilizing approximately $19.3$ million fewer parameters ($180.45$M vs. $199.79$M). The primary sources of the increased computational demand are: the tri-input framework, which necessitates an additional backbone and a supplementary pathway through the encoder, and the integration of the DepthAnything V2 \cite{yang2024depth} model for monocular depth estimation, which alone accounts for $43$ GFLOPs. Ultimately, this computational compromise represents a deliberate  architectural trade-off through the explicit modeling of depth which provides critical geometric priors that justify the higher FLOPs count.

\section{Ablation Studies}

\subsection{The effect of LLIE on Semantic Segmentation performance}

The LLIE process conducted before the segmentation in the low-light configuration of VesperMUSE is essential for recovering lost detail and removing the noise. In this section the effect of the enhancement will be thoroughly studied to determine what improvement it brings in the semantic segmentation process  and if some classes can benefit more from LLIE.

\begin{figure}[H]
    \centering
    \begin{minipage}[b]{0.48\linewidth}  
        \centering
        \includegraphics[width=\linewidth]{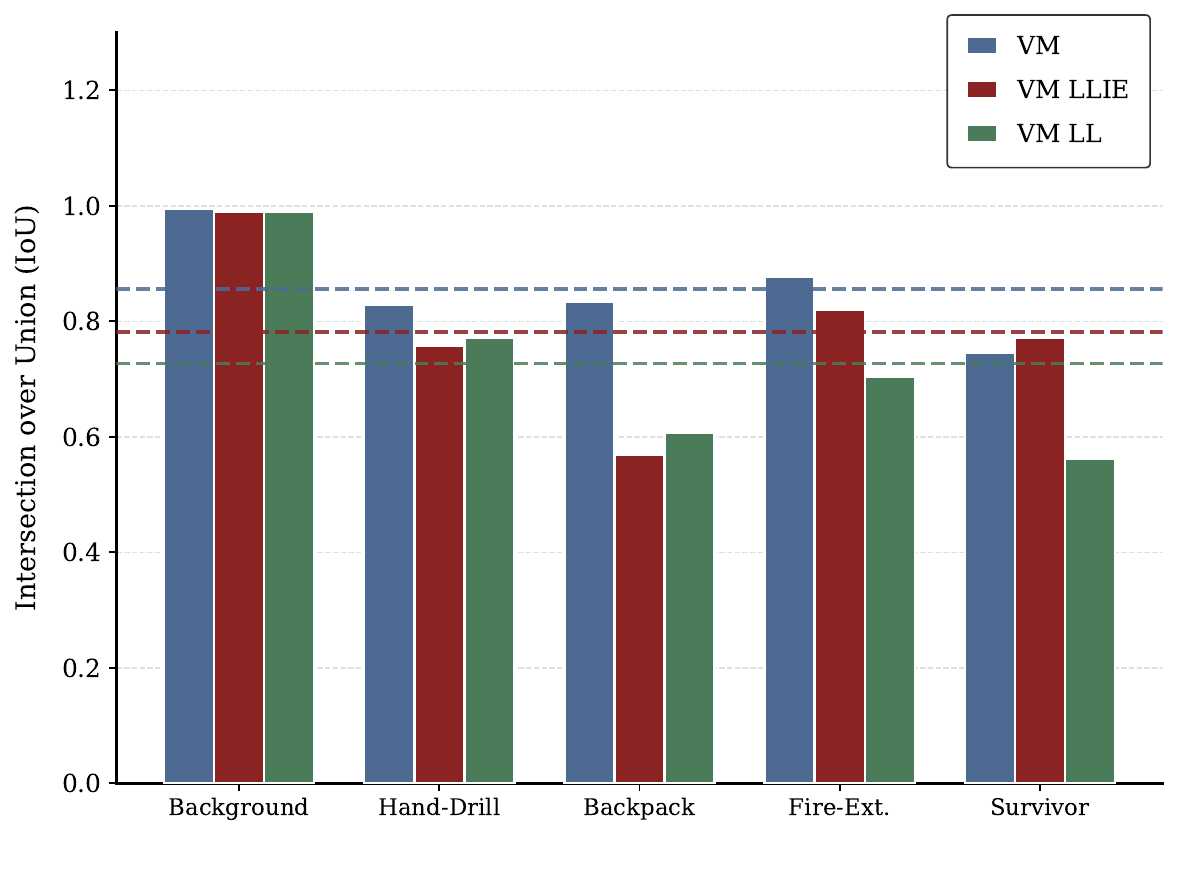}
        \vspace{-2mm}
        \caption{Semantic segmentation performance between different input configuration using: RGB (VM), LLIE RGB (VM LLIE), and low-light RGB without any enhancement (VM LL) on the PST900 dataset \cite{shivakumar2020pst900}.}
        \label{fig:pst900_rgb_ll_llie_comparison_ablation}
    \end{minipage}\hfill
    \begin{minipage}[b]{0.48\linewidth}  
        \centering
        \includegraphics[width=\linewidth]{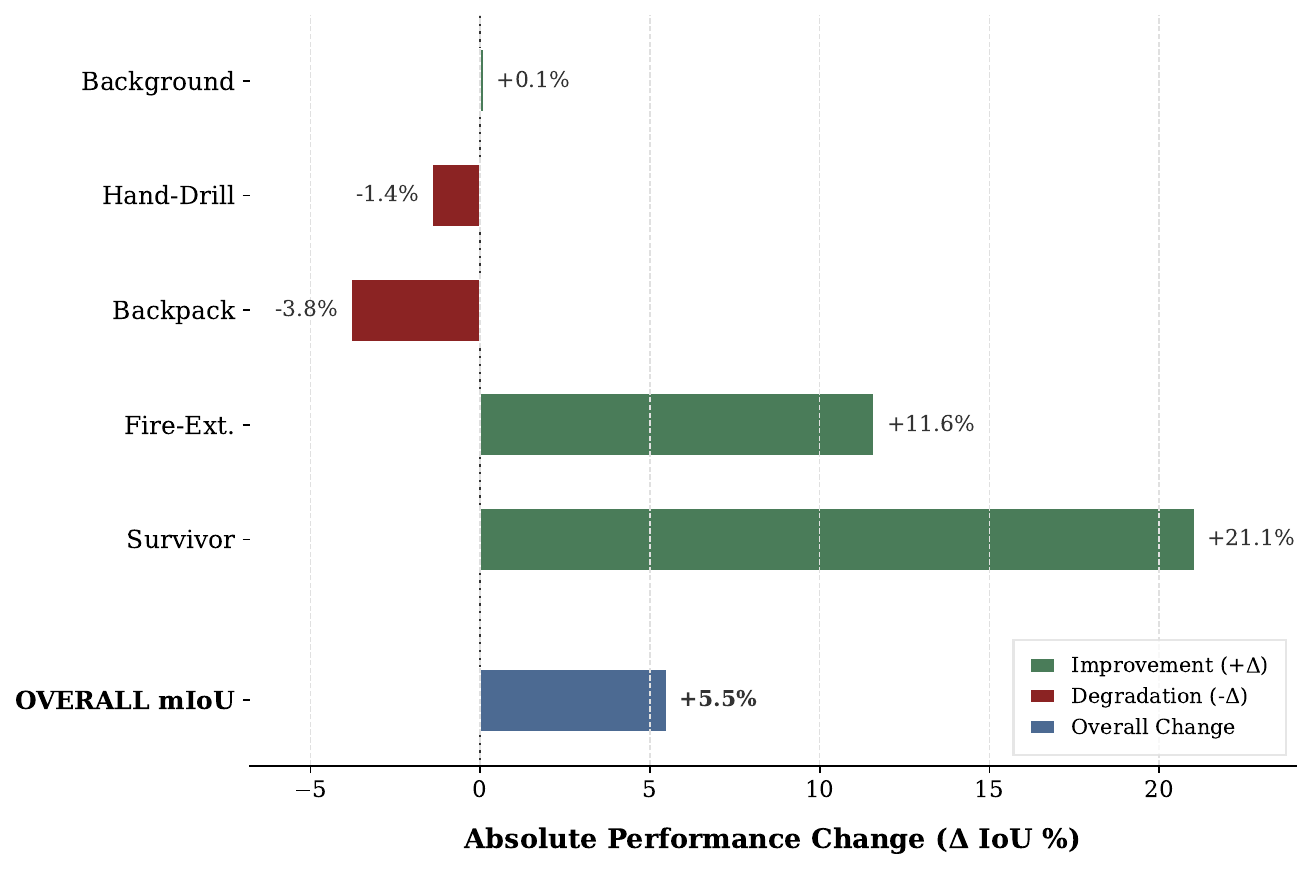}
        \vspace{-2mm}
        \caption{Semantic segmentation performance deltas between VesperMUSE LLIE and VesperMUSE trained and tested on low-light images, highlighting the role of LLIE enhancement on the PST900 dataset \cite{shivakumar2020pst900}.  }
        \label{fig:pst900_rgb_ll_llie_performance_deltas_ablaton}
    \end{minipage}
\end{figure}

\begin{figure}[H]
    \centering
    \begin{minipage}[b]{0.48\linewidth}  
        \centering
        \includegraphics[width=\linewidth]{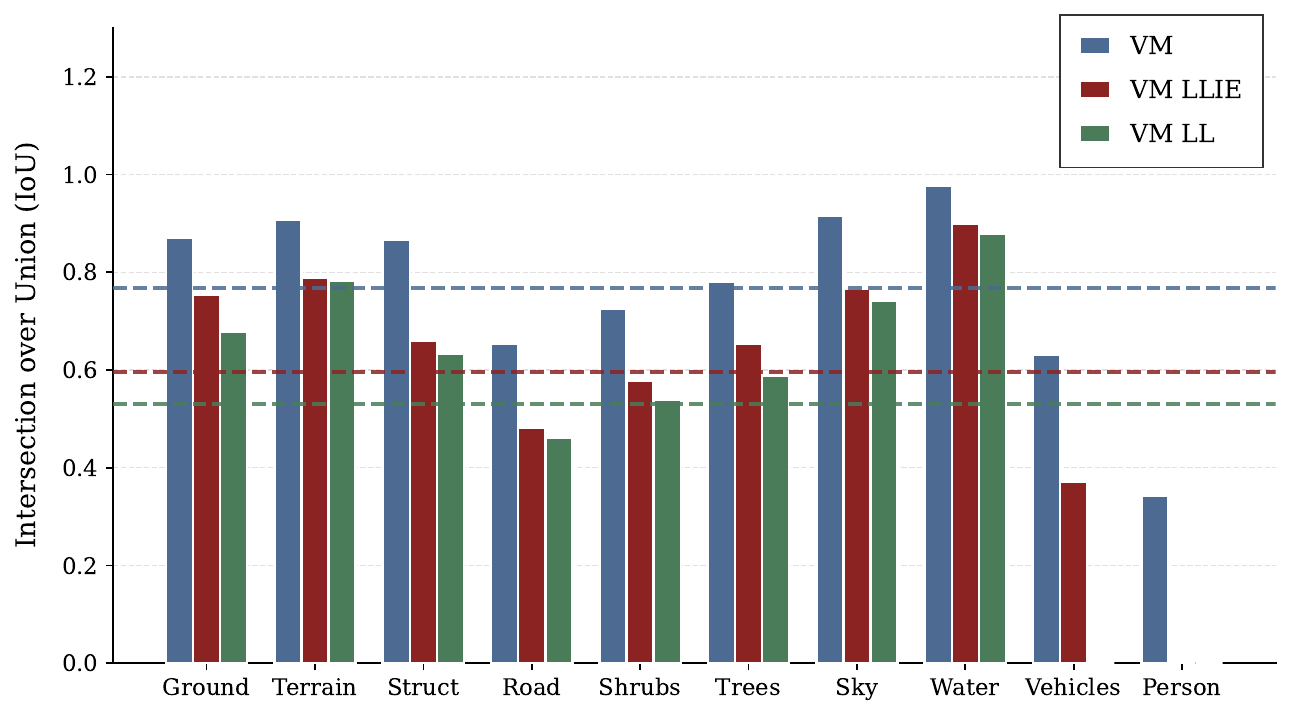}
        \vspace{-2mm}
        \caption{Semantic segmentation performance between different input configuration using: RGB (VM), LLIE RGB (VM LLIE), and low-light RGB without any enhancement (VM LL) on the CART dataset \cite{lee2024cart}.}
        \label{fig:cart_rgb_ll_llie_comparison_ablation}
    \end{minipage}\hfill
    \begin{minipage}[b]{0.48\linewidth}  
        \centering
        \includegraphics[width=\linewidth]{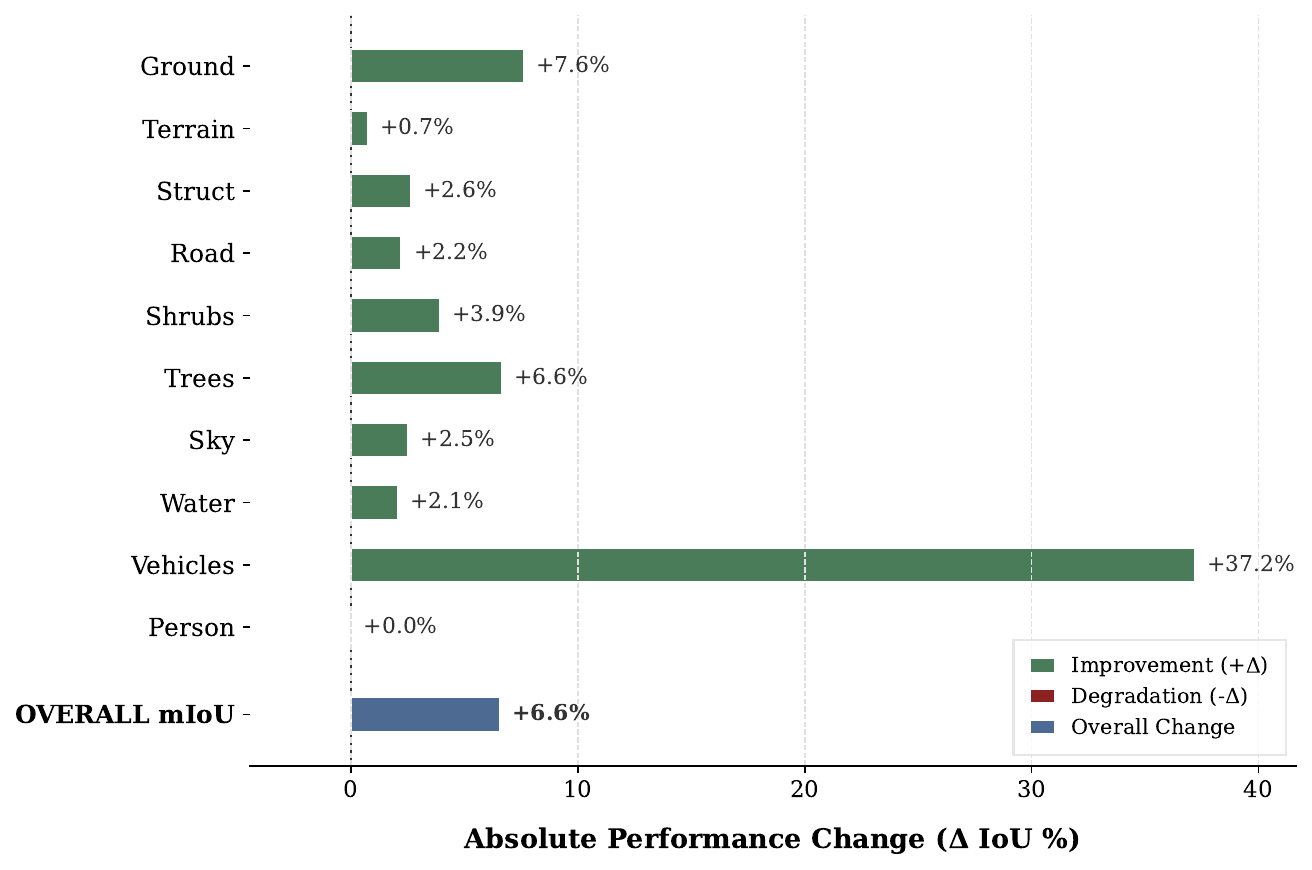}
        \vspace{-2mm}
        \caption{Semantic segmentation performance deltas between VesperMUSE LLIE and VesperMUSE trained and tested on low-light images, highlighting the role of LLIE enhancement on the CART dataset \cite{lee2024cart}.}
        \label{fig:cart_rgb_ll_llie_performance_deltas_ablaton}
    \end{minipage}
\end{figure}

Figures \ref{fig:pst900_rgb_ll_llie_comparison_ablation} and \ref {fig:cart_rgb_ll_llie_comparison_ablation}  show that the RGB configuration of VesperMUSE dominates in terms of performance which is absolutely expected. The main focus point of this study is to see if under a tri-modal input setting with no prior enhancement for the RGB, the performance is decreasing. 

For the PST900 dataset \cite{shivakumar2020pst900} the overall dataset mIoU improvement is strongly influenced by the large performance differences on the \textit{Fire-Ext.} and \textit{Survivor} classes, while the \textit{Hand-Drill} and \textit{Backpack} classes show a slight performance decline. The differences, however, have a considerable offset in the magnitude. While for the \textit{Hand-Drill} and \textit{Backpack} classes it can be considered just a model or training particularity, it is not the same case for \textit{Fire-Ext.} and \textit{Survivor}. A good candidate explanation for this would be the common features shared by the two groups of classes. \textit{Survivor} and \textit{Fire-Ext} are represented using brighter colours and a larger surface covered by their instances, while \textit{Backpack} and \textit{Hand-Drill} often have less contrastive colours and definitely a smaller instance surface.

Conversely, on the CART dataset \cite{lee2024cart}, Figure \ref{fig:cart_rgb_ll_llie_performance_deltas_ablaton} shows a different distribution. Here, the model configuration where the input setting with low-light images which are not enhanced as RGB inputs, provides worse performance for all the classes. The natural light present in CART \cite{lee2024cart} which is distributed more evenly, opposed to the artificial light and the cave environment of PST900 \cite{shivakumar2020pst900}, manages to provide a better colour balance which is preserved across the synthetic low-light adaptation and enhancement process
throughout. The only issue that persists even after the enhancement process has been applied is the absence of the \textit{Person} class among the predictions which can be again attributed, as in the PST900 \cite{shivakumar2020pst900} case, to the small surface covered by noise across its instances.

\subsection{Modality choices and their rank}

The inputs of VesperMUSE can be configured in a large number of ways. Although the standard and recommended configuration is composed of three inputs: RGB, thermal, and surface normals, the model can be run using any number of modalities as inputs with very little changes as the Fusion Self-Attention was specifically designed to structurally support any number of modalities.

The only constraint, however, is that one modality has to be declared as the main modality while the others are auxiliaries. The main modality is used as the main source of information in the encoding stage and also has a larger backbone responsible for extracting the features from it. In all the configurations presented until now, the main modality was the RGB (normal or synthetically low-light adapted and then enhanced), but the purpose of this ablation study is to show how the model behaves under different number of inputs and what are the effects of modifying the rank of the modalities. 

Table \ref{tab:modality_ablation} presents different input configurations and provides evidence for why RGB was chosen as a primary modality even when running under a LLIE configuration and the enhanced images present noise and colour distortions, while the thermal is left unaffected. Furthermore, LLIE RGB is shown to also validate the primary modality advantage when interchangeably switched with the surface normal representations.

\textbf{Thermal and Surface Normals as Auxiliaries. }
A first observation, as shown in Figure \ref{fig:intro_llie_reliability}, is that both modalities bring an important contribution if compared to the segmentation network using only the LLIE RGB  as input. Their impact is very similar when individually employed, with an average of $\approx3\%$ performance deficit when the surface normal is not used and an average of $\approx3.5\%$ performance deficit considering both datasets. The combined benefit is observed mainly on the PST900 \cite{shivakumar2020pst900} dataset with an average improvement of $\approx5\%$  for both modalities, confirming again the fact that the CART dataset \cite{lee2024cart} provides more informativeness in its RGB maps than PST900 \cite{shivakumar2020pst900} because of the different light conditions.

The relevance of the RGB modality is again confirmed when comparing the two configurations where LLIE RGB and surface normal representations are switched so that each one is alternatively a main modality and an auxiliary. The results on both datasets show that there is an average of $\approx7.0\%$ performance deficit when the surface normal map is used as a main modality.

\textbf{Thermal as Main Modality. } 
Given that this ablation study is conducted under low-light configuration, it would be expected for the thermal modality provide more information about the overall composition of the image, as it is not affected by noise, exposure and loss of global and local structures. However, a configuration with thermal used as a main modality, with surface normals, reflectance and illumination as auxiliaries shows a large performance decline compared to the other configurations where LLIE RGB is used as the main modality. 

The reflectance and illuminance maps represent a decomposition of the LLIE RGB image conducted under the assumption of the Retinex Theory \cite{Retinex} which suggests that visual information can be separated into two core components: reflectance responsible for the colour information and illuminance encoding the intensity of the pixels. 

The decomposition can be done using various ways, from classical to neural networks which are trained specifically for accurate representation of both components. However, for the purpose of this experiment, a simpler approach was used, where the illumination is considered as the maximum value across the channels and reflectance as the original image divided by the illumination, both filtered using, a Gaussian Kernel.

\begin{equation}
    L = G_{\sigma} * \left( \max_{c \in \{R,G,B\}} \mathcal{I}_c + \epsilon \right), \qquad R = \frac{I}{G_{\sigma} * \left( \max_{c \in \{R,G,B\}} \mathcal{I}_c + \epsilon \right)}.
\end{equation}

The quantitative results from Table \ref{tab:modality_ablation} confirm that the thermal modality cannot be successfully employed as a main modality, implying that colour information, even if it is shifted by the low-light adaptation and enhancement, provides more important features for the performance of semantic segmentation than the clean thermal imaging.

\begin{table}[ht]
\centering
\setlength{\tabcolsep}{12pt} 
\resizebox{0.7\linewidth}{!}{%
\begin{tabular}{@{}ll cc@{}} 
\toprule
\multirow{2}{*}{\textbf{Primary Modality}} & \multirow{2}{*}{\textbf{Auxiliary Modalities}} & \multicolumn{2}{c}{\textbf{mIoU (\%) $\uparrow$}} \\
\cmidrule(l){3-4} 
& & \textbf{PST900} & \textbf{CART} \\
\midrule
LLIE RGB      & Thermal, Surface Norm.      & \textbf{78.18} & \textbf{59.57} \\
              & Thermal                     & 74.53 & 58.56 \\
              & Surface Norm.               & 73.80 & 58.62 \\
\midrule
Thermal       & Surface Norm., Ref., Illum. & 60.60 & 50.76 \\
\midrule
Surface Norm. & LLIE RGB                    & 67.70 & 55.90 \\
\bottomrule
\end{tabular}%
}
\vspace{0.5em}
\caption{Table analyzing the impact of different primary and auxiliary input modalities on the PST900 \cite{shivakumar2020pst900} and CART \cite{lee2024cart} datasets. Grouping by the primary backbone highlights the performance delta when injecting various auxiliary streams.}
\label{tab:modality_ablation}
\end{table}

\subsection{Architectural motivation}
\textbf{Impact of the Fusion Strategy.} To motivate the design of the proposed Fusion Encoder Block, it is compared against a baseline strategy that relies on naive concatenation. In this baseline, multi-modal features are simply concatenated along the channel dimension prior to passing through a standard Transformer block. As shown in Table \ref{tab:ablation_fusion_decoder}, the proposed cross-attention fusion approach yields a substantial performance improvement, outperforming the concatenation baseline by $\approx 4\%$ mIoU. 

This performance gap highlights the limitations of early concatenation. While concatenation aggregates multi-modal data into a unified tensor, it forces the subsequent layers to implicitly learn the cross-modal alignments from scratch without any structured guidance. In contrast, the proposed fusion mechanism explicitly models the spatial and semantic correlations between the primary and auxiliary modalities. By utilizing cross-attention, the network dynamically queries the auxiliary streams to enhance the primary features, ensuring that the model selectively attends to relevant multi-modal cues rather than treating all concatenated channels with equal initial weight.

\textbf{Replacing Decoders with Convolutions.} As detailed in Section \ref{sec:decoder}, the core feature refinement component within the VesperMUSE decoder is a lightweight, attention-based Transformer block. This module is specifically designed to refine the multi-scale fused features, utilizing self-attention to dynamically weigh and prioritize spatial information that is most relevant to the current semantic context. To validate this design choice, an ablation study was conducted where these Transformer blocks were replaced with standard pointwise convolutions, effectively removing the global context-aggregation capabilities of the attention mechanism while maintaining a similar channel dimensionality. 

    Table \ref{tab:ablation_fusion_decoder}, shows that reverting to a purely convolutional decoder results in a measurable performance degradation, dropping the mIoU from $0.7818$ to $0.7351$ on PST900 \cite{shivakumar2020pst900}, and from $0.5957$ to $0.5320$ on CART \cite{lee2024cart}. While the convolutional substitution slightly reduces the total parameter count (from 180M to 179M), it increases the computational overhead (from 321G to 326G FLOPs). More importantly, the drop in segmentation accuracy confirms that the spatially-aware feature refinement provided by the Transformer block is crucial. The self-attention mechanism allows the network to effectively resolve local ambiguities by leveraging long-range dependencies, a capability that simple pointwise convolutions lack.

\begin{table}[ht]
\centering
\setlength{\tabcolsep}{8pt} 
\resizebox{0.7\linewidth}{!}{%
\begin{tabular}{@{}l cc cc@{}} 
\toprule
\multirow{2}{*}{\textbf{Method}} & \multirow{2}{*}{\textbf{Params (M) $\downarrow$}} & \multirow{2}{*}{\textbf{FLOPs (G) $\downarrow$}} & \multicolumn{2}{c}{\textbf{mIoU (\%) $\uparrow$}} \\
\cmidrule(l){4-5} 
& & & \textbf{PST900} & \textbf{CART} \\
\midrule
VesperMUSE LLIE              & 180 & \textbf{321} & \textbf{78.18} & \textbf{59.57} \\
Concatenation-based Fusion   & 192 & \textbf{321} & 75.42 & 54.44 \\
Convolution Decoder          & \textbf{179} & 326 & 73.51 & 53.20 \\
\bottomrule
\end{tabular}%
}
\vspace{0.5em}
\caption{Ablation study analyzing the impact of different fusion and decoder strategies on the PST900 \cite{shivakumar2020pst900} and CART \cite{lee2024cart} datasets, including parameter counts and computational complexity.}
\label{tab:ablation_fusion_decoder}
\end{table}

\chapter{Conclusions}
\newpage
\section{Known Limitations}

\begin{figure}[t]
    \centering
    \includegraphics[width=\linewidth]{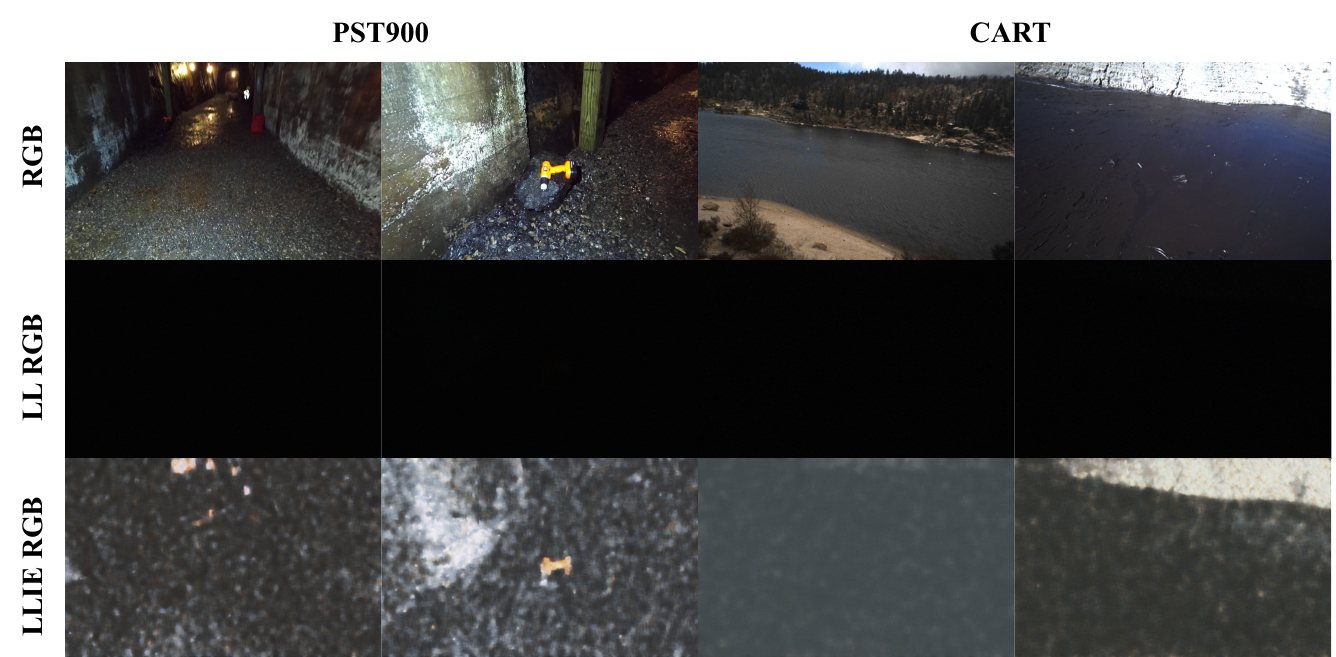}
    \vspace{-2mm} 
    \caption{Qualitative maps highlighting the noise and excessive degradation observed in few synthetically low-light adapted images.}
    \label{fig:pst900_syn_ll_limitations}
\end{figure}

\textbf{Synthetic Low-Light Adaptation Algorithm Strong Degradation. } In the current configuration, the main issue that is persistent within the low-light synthetic pipeline is that a small sample of the low-light samples are excessively degraded so that the LLIE module cannot recover lost information. The source of this issue remains unknown, however it is suspected that the issue comes from increasing the variability of illumination in the low-light conditions. As a result, a very few number of samples from the final light distribution are affected by this issue. Figure \ref{fig:pst900_syn_ll_limitations} presents a part of the affected samples with some managing to capture the overall structure of the image keeping some of the objects observable while others represent a totally noise and artifacts-covered representation which cannot be correctly evaluated by the   segmentation module.

A solution to this problem would be either removing the affected samples or further tuning the synthetic low-light adaptation algorithm to diagnose problematic scenarios and increase the illumination threshold while keeping the noise levels lower than for other samples.

\begin{figure}[t]
    \centering
    \begin{minipage}[b]{0.48\linewidth}
        \centering
        \includegraphics[width=\linewidth]{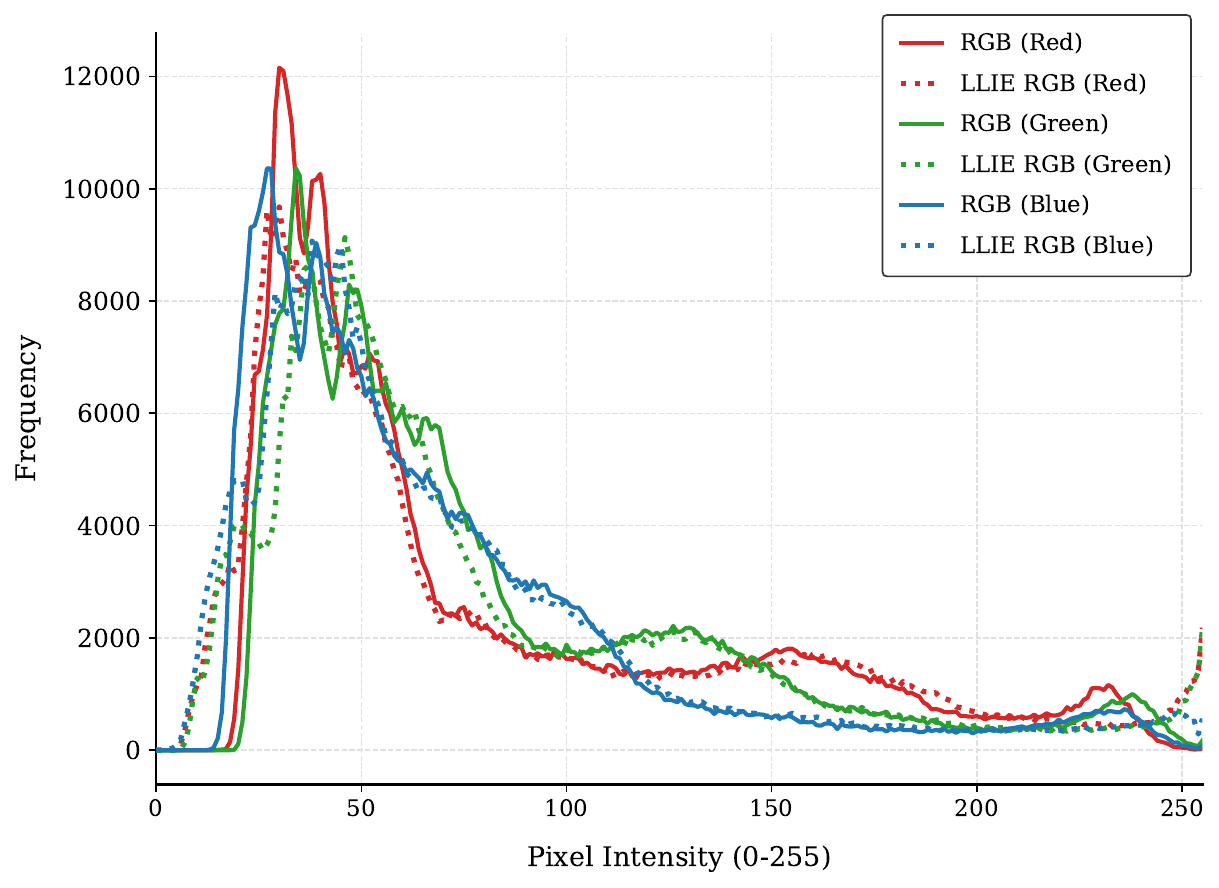}
    \end{minipage}\hfill
    \begin{minipage}[b]{0.48\linewidth}
        \centering
        \includegraphics[width=\linewidth]{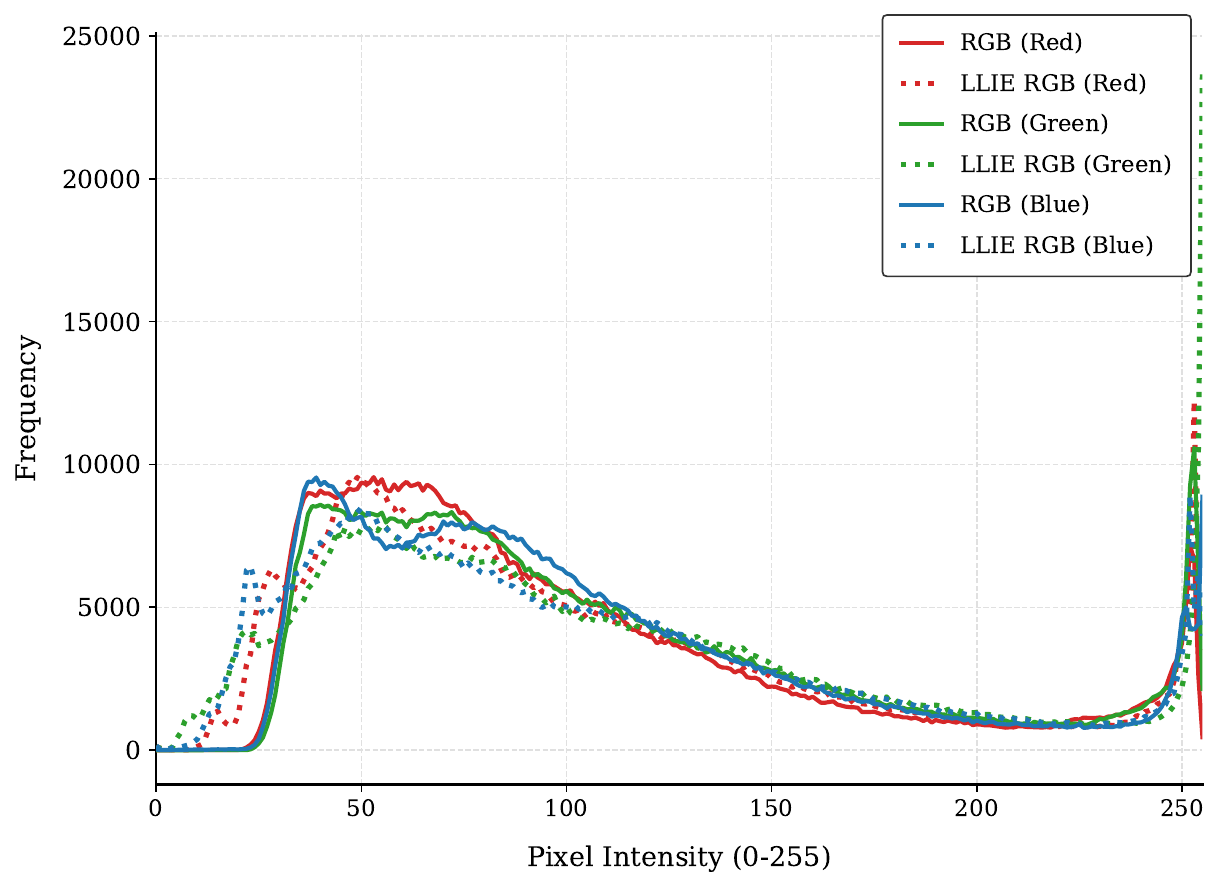}
    \end{minipage}
    \caption{Histogram comparison between the two pairs of RGB and LLIE RGB maps showing the colour balance.}
    \label{fig:limitations_histogram_pair}
\end{figure}

\textbf{LLIE module Output Colour Imbalance. } A comparative analysis of the global intensity distributions (Figure \ref{fig:limitations_histogram_pair}) reveals that the LLIE module introduces luminance-dependent chromatic aberrations via non-linear channel uncoupling. Specifically, in the shadow and mid-tone regimes ($0 \leq I \leq 150$), the model disproportionately attenuates the blue channel relative to the source distribution, inducing an unnatural warm colour cast. Conversely, in the extreme highlight regime ($I \to 255$), the output exhibits anomalous green-channel clipping, with saturation peaks significantly exceeding both the original image limits and concurrent RGB channels. This divergent colour distortion fundamentally disrupts the intrinsic RGB ratios across the dynamic range, highlighting a deficiency in colour constancy and necessitating the integration of cross-channel coupled tone mapping.

Since the current loss function configuration does not have a colour loss integrated, this would strongly benefit the reduction of channel-based colour discrepancies between the well-lit images and the predicted maps from the LLIE module. In this case, the loss function would receive a new term, namely a colour-focused loss.
    
Standard spatial losses penalize the necessary brightness enhancements in low-light processing. To decouple intensity from colour,  RGB channels for each pixel are treated as a 3D vector. By computing the cosine similarity between the input vector ($\mathbf{v}^{in}_i$) and output vector ($\mathbf{v}^{out}_i$) across all $N$ pixels, the colour loss ($\mathcal{L}_{colour}$) selectively penalizes chromatic angular deviation while ignoring magnitude shifts, preserving intrinsic RGB ratios.

\begin{equation}
\mathcal{L}_{colour} = \frac{1}{N} \sum_{i=1}^{N} \left( 1 - \frac{\mathbf{v}^{in}_i \cdot \mathbf{v}^{out}_i}{\|\mathbf{v}^{in}_i\|_2 \|\mathbf{v}^{out}_i\|_2} \right).
\end{equation}

Another solution would be the processing of the input in a different colour space. Operating within an orthogonal colour space like YCbCr inherently avoids cross-channel distortion. An initial transformation ($\Gamma$) separates the signal, allowing the enhancement function to target exclusively the input luminance. Because the original chrominance channels ($C_{b,in}, C_{r,in}$) remain strictly untouched, projecting the composite signal back via the inverse mapping ($\Gamma^{-1}$) generates enhanced RGB outputs ($R_{out}, G_{out}, B_{out}$) free from luminance-driven colour artifacts.

\begin{equation}
\begin{bmatrix} R_{out} \\ G_{out} \\ B_{out} \end{bmatrix} = \Gamma^{-1} \left( \begin{bmatrix} \mathcal{F}(Y_{in}) \\ C_{b,in} \\ C_{r,in} \end{bmatrix} \right).
\end{equation}

    \textbf{Fusion Vision Transformer Computational Requirements. } Currently, as the focus of this work was not directly placed on computational complexity, the computational requirements needed to perform training and inference are high. An analysis of the comparative performance plot reveals distinct trade-offs between model accuracy and computational complexity. VesperMUSE demonstrates the highest performance among the evaluated models, achieving a peak mIoU of $0.7670$ on the CART dataset \cite{lee2024cart}. However, this superior accuracy incurs a higher computational cost relative to the baselines. Conversely, CM-SSM \cite{guo2025cross} emerges as a highly efficient alternative, delivering competitive accuracy (mIoU $\approx$ $0.7460$) while maintaining a significantly lower computational footprint. Overall, while VesperMUSE establishes a new performance ceiling and strictly dominates heavier baselines like MCNet-T \cite{guo2025transferring}, models such as CM-SSM \cite{guo2025cross} provide an optimal balance for resource-constrained applications.

\begin{figure}[H]
    \centering
    \includegraphics[width=0.7\linewidth]{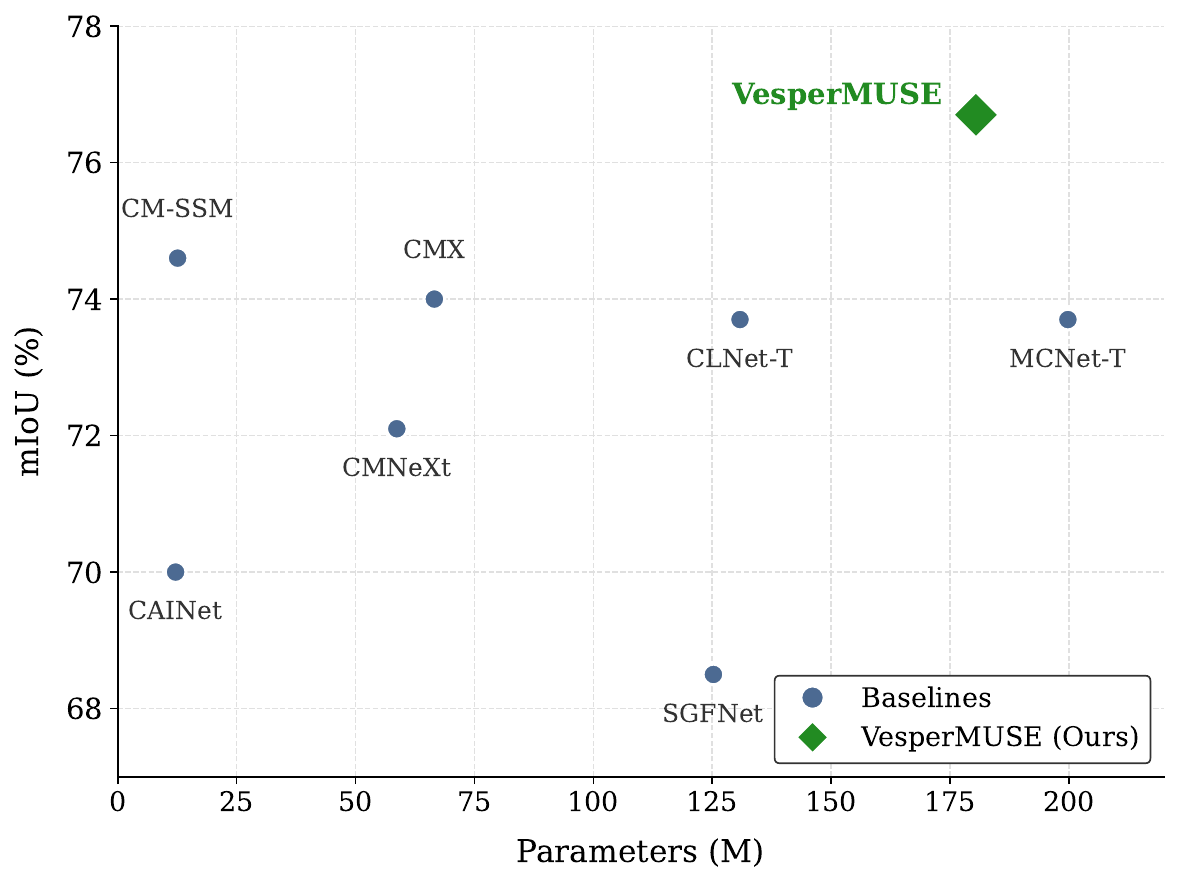}
    \vspace{-2mm} 
    \caption{Visual representation of the ratio between mIoU and the parameter count (M) of SOTA models.}
    \label{fig:miou_params_ratio}
\end{figure}

Besides architectural changes such as backbone change, lowering the projections sizes in the decoder path of the Fusion Vision Transformer or even modality-based pruning (only if two or more features have a high feature similarity), other inference methods like mixed precision (fp16), quantisation (by reducing the precision of the weights to a smaller container like int8/int16) can be taken into account for both the LLIE and semantic segmentation modules.

\section{Future Work}
The subsequent development of VesperMUSE should primarily target both performance enhancement and the reduction of computational complexity. At present, further improvements in performance can be facilitated by adopting more efficient backbone architectures, potentially surpassing those derived from the ResNet family. Additionally, the utilization of modality-specific backbones for thermal and structural inputs represents a promising direction, as these are expected to outperform the current convolutional adaptation strategy employed within the model.

Furthermore, it is worthwhile to investigate alternative representations of image information, such as operating in different colour spaces or explicitly decomposing the input into low- and high-frequency components within the LLIE module. Such approaches may enable a more principled treatment of light irradiation, which has been shown to be an effective strategy in LLIE tasks.

\section{Final Thoughts}
The proposed framework is intended to establish a foundation for future advancements in semantic segmentation under challenging and highly variable illumination conditions. Despite recent progress, the field continues to be constrained by the limited availability of diverse and representative datasets, which significantly hinders the development of robust and generalizable methods.

The scenarios explored in this study reflect realistic and often critical environments, including safety-sensitive missions where reliability is paramount. In such contexts, the deployment of models that are not only performant but also transparent with respect to their limitations becomes essential. This is particularly relevant for applications such as survivor identification, where the consequences of failure are profound. Therefore, advancing DL-based approaches in this domain is not solely a matter of improving accuracy, but also of fostering trust, robustness, and accountability in systems designed to operate under uncertainty.

\clearpage
\addcontentsline{toc}{chapter}{References}
\bibliographystyle{IEEEtran}
\bibliography{main}

\end{document}